\pdfoutput=1

\documentclass[11pt]{article}

\usepackage[]{acl2023}

\usepackage{times}
\usepackage{latexsym}
\usepackage[normalem]{ulem}
\usepackage{booktabs}
\usepackage{amsmath}
\usepackage{multirow}
\usepackage{multicol}
\usepackage{graphicx}
\usepackage{subcaption}
\usepackage{caption}
\usepackage[noabbrev]{cleveref}
\usepackage{subcaption}
\usepackage[T1]{fontenc}

\usepackage[utf8]{inputenc}

\usepackage{microtype}
\newcommand{\nospacetext}[1]{\makebox[0pt][l]{#1}}

\title{Easy to Decide, Hard to Agree:\\ Reducing Disagreements Between Saliency Methods} 

\author{Josip Jukić$^{\clubsuit,1}$, \bf Martin Tutek$^{\clubsuit,\heartsuit,2}$, Jan Šnajder$^{1}$ \\
  $^1$TakeLab, University of Zagreb \hspace{0.5em} $^2$UKP Lab, Technical University of Darmstadt\\
  \texttt{\{josip.jukic, jan.snajder\}@fer.hr}\\
  \texttt{tutek@ukp.informatik.tu-darmstadt.de}
}
\begin{document}
\maketitle

\begin{abstract}
A popular approach to unveiling the black box of neural NLP models is to leverage saliency methods, which assign scalar importance scores to each input component. A common practice for evaluating whether an interpretability method is \textit{faithful} has been to use evaluation-by-agreement -- if multiple methods agree on an explanation, its credibility increases. However, recent work has found that saliency methods exhibit weak rank correlations even when applied to the same model instance and advocated for alternative diagnostic methods. In our work, we demonstrate that rank correlation is not a good fit for evaluating agreement and argue that Pearson-$r$ is a better-suited alternative. We further show that regularization techniques that increase faithfulness of attention explanations also increase agreement between saliency methods. By connecting our findings to instance categories based on training dynamics, we show that the agreement of saliency method explanations is very low for easy-to-learn instances. Finally, we connect the improvement in agreement across instance categories to local representation space statistics of instances, paving the way for work on analyzing which intrinsic model properties improve their predisposition to interpretability methods.
\end{abstract}

\section{Introduction}
\label{sec:intro}

Following the meteoric rise of the popularity of neural NLP models during the neural revolution, they have found practical usage across a plethora of domains and tasks.
However, in a number of high-stakes domains such as law \citep{kehl2017algorithms}, finance \citep{grath2018interpretable}, and medicine \citep{caruana2015intelligible}, the opacity of deep learning methods needs to be addressed.
In the area of explainable artificial intelligence (XAI), one of the major recent efforts is to unveil the neural black box and produce explanations for the end-user.
There are various approaches to rationalizing model predictions, such as using the attention mechanism \cite{bahdanau2014neural}, saliency methods \cite{denil2014extraction,bach2015pixel,ribeiro2016should,lundberg2016unexpected,shrikumar2017learning,sundararajan2017axiomatic}, rationale generation by-design \cite{lei-etal-2016-rationalizing,bastings2019interpretable,jain2020learning}, or self-rationalizing models \cite{marasovic2021few}.
These methods have to simultaneously satisfy numerous desiderata to have practical application in high-stakes scenarios: they have to be \textit{faithful} -- an accurate representation of the inner reasoning process of the model, and \textit{plausible} -- convincing to human stakeholders.

When evaluating faithfulness in using attention as explanations, \citet{jain2019attention} have shown that attention importance scores do not correlate well with gradient-based measures of feature importance.
The authors state that although gradient-based measures of feature importance should not be taken as ground truth, one would still expect importance measures to be highly agreeable, bringing forth the \textit{agrement-as-evaluation} paradigm \cite{abnar2020quantifying,meister2021sparse}.
While the imperfect agreement is something one could expect as interpretability methods differ in their formulation, and it is reasonable to observe differences in importance scores, subsequent work has shown that saliency methods exhibit low agreement scores even when applied to the \textit{same model instance} \cite{neely2021order}.
Since a single trained model instance can only have a single feature importance ranking for its decision, disagreement of saliency methods implies that at least one, if not all methods, do not produce faithful explanations -- placing doubt on their practical relevance.
It has been hypothesized that unfaithfulness of attention is caused by input entanglement in the hidden space \cite{jain2019attention}.
This claim has later been experimentally verified through results showing that regularization techniques targeted to reduce entanglement significantly improve the faithfulness of attention-based explanations \cite{mohankumar2020towards,tutek2020staying}.
While entanglement in the hidden space is clearly a problem in the case of attention explanations, where attention weights directly pertain to hidden states, we also hypothesize that representation entanglement could cause similar issues for gradient- and propagation-based explainability methods -- which might not be able to adequately disentangle importance when propagating toward the inputs.

In our work, we first take a closer look at whether the rank correlation is an appropriate method for evaluating agreement and confirm that, as hypothesized in previous work, small differences in values of saliency scores significantly affect agreement scores.
We argue that a linear correlation method such as Pearson-$r$ is a better-motivated choice since the exact ranking order of features is not as crucial for agreement as the relative importance values, which Pearson-$r$ adequately captures.
We hypothesize that the cause of saliency method disagreements is rooted in representation entanglement and experimentally show that agreement can be significantly improved by regularization techniques such as tying \cite{tutek2020staying} and conicity \cite{mohankumar2020towards}.
The fact that regularization methods, which were originally aimed at improving faithfulness of attention, also improve agreement between saliency methods suggests that the two problems have the same underlying cause.
Taking the analysis deeper, we apply techniques from dataset cartography \cite{swayamdipta2020dataset} and show that, surprisingly, the explanations of easy-to-learn instances exhibit a lower agreement than of ambiguous instances.
We further analyze how local curvature of the representation space morphs when regularization techniques are applied, paving the way for further analysis of (dis)agreements between interpretability methods. 

\section{Background and Related Work}
\label{sec:rw}

Explainability methods come in different flavors determined by the method of computing feature importance scores.
Saliency methods perform \textit{post-hoc} analysis of the trained black-box model by either leveraging gradient information \cite{denil2014extraction,sundararajan2017axiomatic}, modifying the backpropagation rules \cite{bach2015pixel,shrikumar2017learning}, or training a shallow interpretable model to locally approximate behavior of the black-box model \cite{ribeiro2016should}, all with the goal of assigning scalar saliency scores to input features.
Alternatively, if the analyzed model is capable of generating text, one can resort to self-rationalization by prompting the trained model to generate an explanation for its decision \cite{marasovic2021few}.
In contrast to \textit{post-hoc} explanations, \textit{inherently interpretable} models produce explanations as part of their decision process, either by masking a proportion of input tokens and then performing prediction based on the remaining \textit{rationale} \cite{lei-etal-2016-rationalizing,bastings2019interpretable,jain2020learning}, or jointly performing prediction and rationale generation in cases where datasets with annotated rationales are available \cite{camburu2018snli}.
For some time, the attention mechanism \cite{bahdanau2014neural} has also been considered inherently interpretable. However, the jury is still out on whether such explanations can be considered faithful  \cite{jain2019attention,wiegreffe2019attention,tutek2020staying,bastings2020elephant}.

%
%


\textit{Faithfulness} is one of the most important desiderata of explanation methods \cite{jacovi-goldberg-2020-towards} -- faithful explanations are those that are true to the inner decision-making process of the model.
Approaches to evaluating faithfulness rely on measuring how the confidence of the model changes when inputs are perturbed \cite{kindermans2019reliability} or completely dropped from the model \cite{li2016understanding,serrano2019attention}.
However, perturbations to input often result in corrupted instances that fall off the data manifold and appear nonsensical to humans \cite{feng2018pathologies} or fail to identify all salient tokens properly \cite[ROAR;][]{hooker2019benchmark} -- raising questions about the validity of perturbation-based evaluation.
Recursive ROAR \cite{madsen2021evaluating} alleviates the issues of its predecessor at the cost of requiring many prohibitively expensive retraining steps, further motivating us to seek efficient solutions which do not require retraining the model multiple times.
Another option is to leverage the \textit{evaluation-by-agreement} \cite{jain2019attention} paradigm, which states that an interpretability method should be highly agreeable with other methods to be considered faithful.
However, since empirical evidence has shown that saliency methods exhibit poor agreement between their explanations \cite{neely2021order}, \citet{atanasova2020diagnostic} recommend practitioners consider alternative methods for evaluating the quality of interpretability methods, such as diagnostic tests.
Finally, methods such as data staining \cite{sippy2020data} and lexical shortcuts \cite{bastings2021will} artificially introduce tokens that act as triggers for certain classes -- creating a ground truth for faithfulness which can be used as a comparison.
Nevertheless, such methods have a certain drawback in that they only offer the ground truth importance of a few artificially inserted tokens, but offer no insight regarding the relative importance of the remainder of the input.
Each of the aforementioned methods for estimating faithfulness of interpretability methods has its drawbacks \cite{jacovi-goldberg-2020-towards}, and we argue each should be taken in conjunction with others to increase the credibility of their collective verdict. 
\section{Preliminaries}

In this section, we delineate our experimental setup, detailing the considered datasets, models, their training procedure, the saliency methods which we use to interpret the decisions of the models, and the regularization techniques we use to improve agreement between saliency methods.

\subsection{Datasets}
\label{subsec:datasets}
Leaning on the work of \citet{neely2021order}, which motivated us to explore the valley of explainability, we aim to investigate the protruding problem of low agreement between saliency methods. We investigate three different types of single-sequence binary classification tasks on a total of four datasets. In particular, we evaluate sentiment classification on the movie reviews \cite[\textbf{IMDB};][]{maas-etal-2011-learning} and the Stanford Sentiment Treebank \cite[\textbf{SST-2};][]{socher-etal-2013-parsing} datasets, using the same data splits as \citet{jain2019attention}. We include two more tasks, examining the subjectivity dataset \cite[\textbf{SUBJ};][]{pang-lee-2004-sentimental}, which classifies movie snippets into subjective or objective, and question type classification  \cite[\textbf{TREC};][]{li-roth-2002-learning}. To frame the TREC task as binary classification, we select only the examples labeled with the two most frequent classes (ENTY -- entities, HUM -- human beings) and discard the rest.

\subsection{Models}
For comparability, we opt for the same models as \citet{neely2021order}. Specifically, we employ the Bi-LSTM with additive self-attention \cite[\textbf{\textsc{jwa}};][]{jain2019attention}.
We initialize word representations for the \textsc{jwa} model to $300$-d GloVe embeddings  \cite{pennington-etal-2014-glove}. 
We also employ a representative model from the Transformer family  \cite{vaswani-etal-2017-attention} in DistilBERT  \cite[\textbf{\textsc{dbert}};][]{sanh-etal-2019-distilbert}.

Both models work similarly: the input sequence of tokens $\{x_1, \ldots, x_T\}$ is first embedded $\{e_1, \ldots, e_T\}$ and then contextualized $\{h_1, \ldots, h_T\}$ by virtue of an LSTM network or a Transformer.
The sequence of contextualized hidden states is then aggregated to a sequence representation $h$, which is then fed as input to a decoder network. 

\subsection{Explainability Methods}
We make use of ready-made explainability methods from the propagation- and gradient-based families used by \citet{neely2021order}: Deep-LIFT \cite{shrikumar2017learning}, Integrated Gradients \cite[Int-Grad;][]{sundararajan2017axiomatic} and their Shapley variants \cite{lundberg2016unexpected}, Deep-SHAP and Grad-SHAP.\footnote{We use implementations of explainability methods from the Captum framework: \url{https://github.com/pytorch/captum}}
Since we evaluate agreement on the entire test set instead of an instance subset \cite{neely2021order}, we exclude LIME \cite{ribeiro2016should} from the comparison as it is not computationally feasible to train the surrogate model for all test instances across all training setups.

Each saliency method produces a set of importance scores for each input (sub)word token.
When evaluating the agreement between different saliency methods for a single trained model, one would expect the importance scores for the same input instance to be similar, as the same set of parameters should produce a unique and consistent importance ranking of input tokens.

\subsection{Regularization Methods}

As alluded to earlier, we suspect one cause of disagreement between saliency method explanations to be rooted in representation entanglement.
To counteract this issue, we employ two regularization schemes that have been shown to improve the faithfulness of the attention mechanism as a method of interpretability: \textsc{conicity} \cite{mohankumar2020towards} and \textsc{tying} \cite{tutek2020staying}.
Both of these methods address what we believe is the same underlying issue in \textit{recurrent} models -- the fact that hidden representations $h_t$ are often very similar to each other, indicating that they act more as a sequence representation rather than a contextualization of the corresponding input token $x_t$.

Each regularization method tackles this problem in a different manner. \textsc{conicity} aims to increase the angle between each hidden representation and the mean of the hidden representations of a single instance.
The authors first define the \textit{alignment to mean} (ATM) for each hidden representation as the cosine similarity of that representation to the average representation:
\begin{equation}
    \text{ATM}(h_i, \mathbf{H}) = \text{cosine}(h_i, \frac{1}{T}\sum_{j=1}^T h_j)
    \label{eq:atm}
\end{equation}
\noindent where $\mathbf{H}=\{h_1, \ldots, h_T\}$ is the set of hidden representations for an instance of length $T$. Conicity is then defined as the average ATM for all hidden states $h_i \in H$:
\begin{equation}
    \text{conicity}(\mathbf{H}) = \frac{1}{T}\sum_{i=1}^T \text{ATM}(h_i, H)
    \label{eq:conicity}
\end{equation}
A conicity value implies that all hidden representations exist in a narrow cone and have high similarity -- to counteract this unwanted effect, during training, we minimize this regularization term weighted by $\lambda_{con}$ along with the binary cross entropy loss.

Similarly, \textsc{tying} also aims to incentivize differences between hidden states by enforcing them to ``stay true to their word'' through minimizing the $L_2$ norm of the difference between each hidden state $h_t$ and the corresponding input embedding $e_t = \text{embed}(x_t)$:
\begin{equation}
    \text{tying}(\mathbf{H}, \mathbf{E}) = \frac{1}{T} \sum_{i=1}^T \lVert h_i - e_i \rVert_2^2
    \label{eq:tying}
\end{equation}
\noindent where $\mathbf{E} = \{e_1, \ldots, e_T\}$ is the sequence of embedded tokens. During training, we minimize this regularization term weighted by $\lambda_{tying}$ along with the binary cross entropy loss.

By penalizing the difference between hidden representations and input embedding, one achieves two goals: (1) the embedding and hidden state representation spaces become better aligned, and (2) each hidden representation comes closer to its input embedding.
The latter enforces hidden states to differ from each other: because different embeddings represent the semantics of different tokens, their representations should also differ, and this effect is then also evident in the hidden representations.

Although both works introduced other methods of enforcing differences between hidden states, namely orthogonal-LSTM and masked language modeling as an auxiliary task, we opt for \textsc{conicity} and \textsc{tying} as they were both shown to be more efficient and more stable in practice.


\section{Improving Agreement}
\label{sect:impr-agreement}
In this section, we present two modifications of the existing \textit{evaluation-by-agreement} procedure: (1) substituting rank-correlation with a linear correlation measure, which is more robust to rank changes caused by small differences in importance weights, and (2) regularizing the models with the goal of reducing entanglement in the hidden space, and as a consequence, improving agreement. 

\subsection{Choice of Correlation Metric}
\label{sub:corr-metric}
\begin{figure}[t!]
\centering
\includegraphics[width=\linewidth]{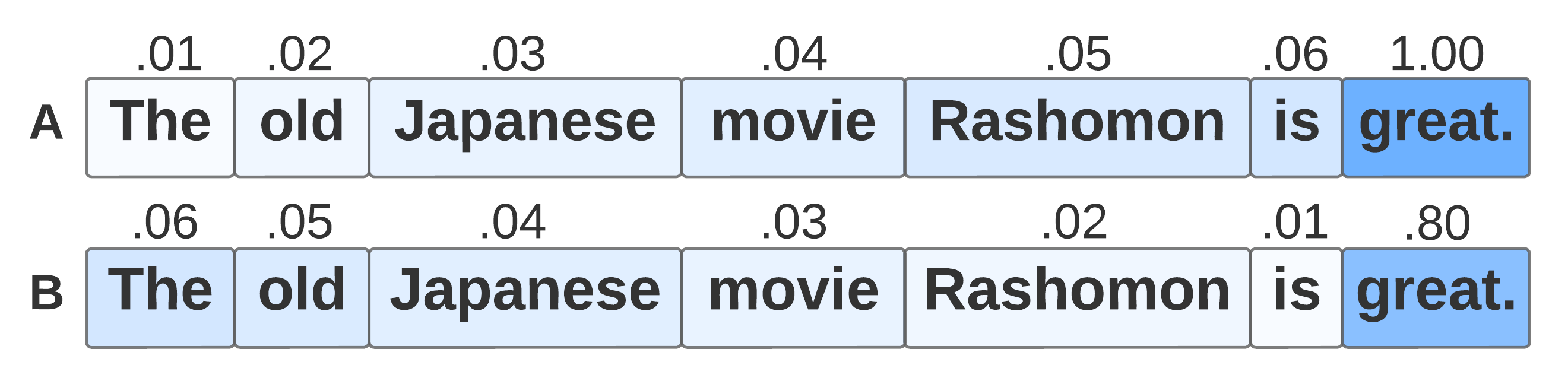}
\caption{A toy example of sentiment classification illustrating the problems with Kendall-$\tau$. The corresponding agreement on the shown example is $k_\tau = -.43$ and $p_r = .99$. Token opacity indicates higher saliency score, i.e., token relevance, written out at the top of the token box. Each of the two explainability methods A and B outputs its saliency scores. The value of Kendall-$\tau$ is much lower than Pearson's correlation, because the irrelevant tokens are perturbed, despite the fact that the tokens are correctly partitioned into more important and less important ones.}
\label{fig:kendall_problem}
\end{figure}

Previous work \cite{jain2019attention,neely2021order} has evaluated the agreement between two explainability methods by using rank-correlation as measured by Kendall-$\tau$ \cite{kendall1938new}.
Although Kendall-$\tau$ is generally more robust than Spearman's rank correlation, i.e., it has smaller gross-error sensitivity \cite{croux2010influence}, we still face difficulties when using Kendall-$\tau$ for evaluating agreement.
As \citet{jain2019attention} also note, perturbations in ranks assigned to tokens in the tail of the saliency distribution have a large influence on the agreement score.
In addition, rankings are also unstable when saliency scores for the most relevant tokens are close to one another.
In \Cref{fig:kendall_problem}, we illustrate the deficiencies of using rank correlation on a toy example of explaining sentiment classification.
While saliency scores attributed to tokens differ slightly, the differences in rank order are significant, lowering agreement according to Kendall-$\tau$ due to the discretization of raw saliency scores when converted into ranks.
We believe that a better approach to approximating agreement is to use a linear correlation metric such as Pearson's $r$, as it evaluates whether both saliency methods assign similar importance scores to the same tokens -- which is a more robust setup if we assume small amounts of noise in importance attribution between different methods.
%
%
\begin{table}[t!]
\begin{subtable}{.48\textwidth}
\centering
\small
\begin{tabular}{lrrrrrrr}
\toprule
& & \multicolumn{2}{c}{D-SHAP} & \multicolumn{2}{c}{G-SHAP} & \multicolumn{2}{c}{Int-Grad} \\
\cmidrule(lr){3-4} \cmidrule(lr){5-6} \cmidrule(lr){7-8}
& & $k_\tau$ & $p_r$ & $k_\tau$ & $p_r$ & $k_\tau$ & $p_r$ \\
\midrule
\multirow{4}{*}{\rotatebox[origin=c]{90}{DeepLIFT}}
& SUBJ & $\mathbf{1.}$ & $\mathbf{1.}$ & $.31$ & $\mathbf{.45}$ & $.43$ & $\mathbf{.64}$ \\
& SST & $\mathbf{1.}$ & $\mathbf{1.}$ & $.30$ & $\mathbf{.47}$ & $.35$  & $\mathbf{.54}$ \\
& TREC & $\mathbf{1.}$ & $\mathbf{1.}$ & $.12$ & $\mathbf{.31}$ & $.15$ & $\mathbf{.33}$ \\
& IMDB & $\mathbf{1.}$  & $\mathbf{1.}$ & $.29$  & $\mathbf{.59}$ & $.28$ & $\mathbf{.60}$ \\
\midrule
\multirow{4}{*}{\rotatebox[origin=c]{90}{D-SHAP}}
& SUBJ & & & $.31$ & $\mathbf{.45}$ & $.43$ & $\mathbf{.64}$ \\
& SST & & & $.30$ & $\mathbf{.47}$ & $.35$ & $\mathbf{.54}$ \\
& TREC & & & $.12$ & $\mathbf{.31}$ & $.15$ & $\mathbf{.33}$ \\
& IMDB & & & $.29$ & $\mathbf{.60}$ & $.28$ & $\mathbf{.60}$ \\
\midrule
\multirow{4}{*}{\rotatebox[origin=c]{90}{G-SHAP}}
& SUBJ & & & & & $.62$ & $\mathbf{.78}$ \\
& SST & & & & & $.70$ & $\mathbf{.87}$ \\
& TREC & & & & & $.66$ & $\mathbf{.85}$ \\
& IMDB & & & & & $.68$ & $\mathbf{.94}$ \\
\bottomrule
\end{tabular}
\subcaption{\textsc{jwa}}
\end{subtable}

\bigskip

\begin{subtable}{.48\textwidth}
\centering
\small
\begin{tabular}{lrrrrrrr}
\toprule
& & \multicolumn{2}{c}{D-SHAP} & \multicolumn{2}{c}{G-SHAP} & \multicolumn{2}{c}{Int-Grad} \\
\cmidrule(lr){3-4} \cmidrule(lr){5-6} \cmidrule(lr){7-8}
& & $k_\tau$ & $p_r$ & $k_\tau$ & $p_r$ & $k_\tau$ & $p_r$ \\
\midrule
\multirow{4}{*}{\rotatebox[origin=c]{90}{DeepLIFT}}
& SUBJ & $.24$ & $\mathbf{.44}$ & $.10$ & $\mathbf{.19}$ & $.12$ & $\mathbf{.21}$ \\
& SST & $.19$ & $\mathbf{.34}$ & $.09$ & $\mathbf{.17}$ & $.10$  & $\mathbf{.20}$ \\
& TREC & $.16$ & $\mathbf{.30}$ & $.12$ & $\mathbf{.25}$ & $.12$ & $\mathbf{.26}$ \\
& IMDB & $.28$  & $\mathbf{.51}$ & $.11$  & $\mathbf{.24}$ & $.13$ & $\mathbf{.27}$ \\
\midrule
\multirow{4}{*}{\rotatebox[origin=c]{90}{D-SHAP}}
& SUBJ & & & $.11$ & $\mathbf{.22}$ & $.13$ & $\mathbf{.24}$ \\
& SST & & & $.10$ & $\mathbf{.19}$ & $.11$ & $\mathbf{.23}$ \\
& TREC & & & $.13$ & $\mathbf{.28}$ & $.14$ & $\mathbf{.30}$ \\
& IMDB & & & $.12$ & $\mathbf{.26}$ & $.14$ & $\mathbf{.30}$ \\
\midrule
\multirow{4}{*}{\rotatebox[origin=c]{90}{G-SHAP}}
& SUBJ & & & & & $.36$ & $\mathbf{.58}$ \\ & SST & & & & & $.31$ & $\mathbf{.54}$ \\
& TREC & & & & & $.42$ & $\mathbf{.71}$ \\ & IMDB & & & & & $.29$ & $\mathbf{.55}$ \\
\bottomrule
\end{tabular}
\subcaption{\textsc{dbert}}
\end{subtable}

\caption{Agreement between pairs of saliency methods in terms of Kendall-$\tau$ ($k_\tau$) and Pearson-$r$ ($p_r$) for \textsc{base} variants of (a) \textsc{jwa} and (b) \textsc{dbert}. We average the agreement over five runs with different seeds. D-SHAP and G-SHAP denote Deep-SHAP and Grad-SHAP, respectively. The values in bold indicate which agreement is higher between the two metrics.}
\label{tab:agreement}

\end{table}

We now evaluate how Pearson's $r$ ($p_r$) compares to Kendall-$\tau$ ($k_\tau$) when evaluating agreement. 
In \Cref{tab:agreement} we compare agreement scores produced by $p_r$ and $k_\tau$ across all datasets for \textsc{jwa} and \textsc{dbert}, respectively. Across all datasets and models, we observe consistently higher agreement values for $p_r$. We take this as an indication that the minor differences between saliency scores on tokens with approximately the same relative importance produced different rankings which were harshly penalized by $k_\tau$. We argue that the higher correlation scores reported by the Pearson correlation coefficient are a better estimate for agreement between saliency scores, and we advocate for its use rather than rank correlation.

\subsection{Regularizing Models}
\label{sub:reg-models}

\begin{table}
\small
\centering
\begin{tabular}{lrrrrrrr}
\toprule
& & \multicolumn{3}{c}{$\tau_k$} & \multicolumn{3}{c}{$r_p$} \\
\cmidrule(lr){3-5} \cmidrule(lr){6-8} \cmidrule(lr){7-8}
& & B & C & T & B & C & T \\
\midrule
\multirow{4}{*}{\rotatebox[origin=c]{90}{\textsc{jwa}}}
& SUBJ & $.52$ & $.48$ & $\mathbf{.65}$\nospacetext{$^\dagger$} & $.66$ & $.70$ & $\mathbf{.88}$\nospacetext{$^\dagger$}  \\
& SST & $.50$ & $.67$ & $\mathbf{.68}$ & $.65$ & $\mathbf{.90}$\nospacetext{$^\dagger$} & $.86$ \\
& TREC & $.37$ & $\mathbf{.77}$\nospacetext{$^\dagger$} & $.68$ & $.52$  & $\mathbf{.98}$\nospacetext{$^\dagger$} & $.93$ \\
& IMDB & $.47$ & $.52$ & $\mathbf{.60}$\nospacetext{$^\dagger$} & $.72$  & $.64$ & $\mathbf{.80}$\nospacetext{$^\dagger$} \\
\midrule
\multirow{4}{*}{\rotatebox[origin=c]{90}{\textsc{dbert}}}
& SUBJ & $.18$ & $.28$ & $\mathbf{.36}$\nospacetext{$^\dagger$} & $.31$ & $.48$ & $\mathbf{.57}$\nospacetext{$^\dagger$} \\
& SST & $.15$ & $.15$ & $\mathbf{.33}$\nospacetext{$^\dagger$} & $.28$ & $.27$ & $\mathbf{.60}$\nospacetext{$^\dagger$} \\
& TREC & $.18$ & $.17$ & $\mathbf{.28}$\nospacetext{$^\dagger$} & $.35$  & $.34$ & $\mathbf{.53}$\nospacetext{$^\dagger$} \\
& IMDB & $.18$ & $.20$ & $\mathbf{.24}$\nospacetext{$^\dagger$} & $.36$ & $.42$ & $\mathbf{.51}$\nospacetext{$^\dagger$} \\
\bottomrule
\end{tabular}
\caption{Average agreement over all pairs of saliency methods. We report agreement in terms of Kendall-$\tau$ ($k_\tau$) and Pearson-$r$ ($r_p$) from the last epoch for three model variants: B -- base, C -- conicity, and T -- tying. \textbf{Bold} numbers indicate highest agreement among the three model flavors. Results are averages over $5$ runs with different seeds. We ran one-sided Wilcoxon signed-rank tests to check for statistical significance. Agreement values significantly higher ($p < .05$) than the other variants in two individual Wilcoxon tests are marked with a $\dagger$. We adjusted the $p$-values for family-wise error rate due to multiple tests using the Holm-Bonferroni method.}
\label{tab:avg_agr}
\end{table}

\begin{table}[t!]
\small
\centering
\begin{tabular}{lrrrr}
\toprule
& & Base & Conicity & Tying \\
\midrule
\multirow{4}{*}{\rotatebox[origin=c]{90}{\textsc{dbert}}}
& SUBJ & $.93_{.01}$ & $.90_{.02}$ & $.93_{.00}$ \\
& SST & $.83_{.00}$ & $.83_{.01}$ & $.82_{.01}$ \\
& TREC & $.92_{.01}$ & $.92_{.01}$ & $.91_{.01}$ \\
& IMDB & $.86_{.01}$ & $.86_{.01}$ & $.88_{.00}$ \\
\midrule
\multirow{4}{*}{\rotatebox[origin=c]{90}{\textsc{jwa}}}
& SUBJ & $.92_{.00}$ & $.90_{.00}$ & $.89_{.00}$ \\
& SST & $.78_{.04}$ & $.76_{.02}$ & $.78_{.02}$ \\
& TREC & $.89_{.02}$ & $.86_{.01}$ & $.89_{.01}$ \\
& IMDB & $.89_{.00}$ & $.88_{.00}$ & $.86_{.00}$ \\
\bottomrule
\end{tabular}
\caption{$F_1$ scores on \textit{test} sets across datasets for \textsc{dbert} and \textsc{jwa}. We report the test results on epochs in which the model had the best performance on the validation set. Columns correspond to the base and regularized models. Numbers in subscript denote standard deviation on $5$ runs with different seeds.}
\label{tab:f1_scores}
\end{table}

Our next goal is to improve agreement between saliency methods through intervention in the training procedure, namely by applying regularization to promote disentanglement in the hidden space. 
In \Cref{tab:avg_agr} we report correlation scores on the test splits of all datasets for regularized models (\textsc{conicity}, \textsc{tying}) and their unregularized variants (\textsc{base}).
We notice that both regularization techniques have a positive effect on agreement across both correlation metrics, indicating that regularization techniques alleviate a deeper issue that also affects the interpretability of attention weights. 
In \Cref{tab:f1_scores} we report $F_1$ scores on the test set for the regularized and unregularized models with the best performance on the validation split.
We observe that regularized models generally perform comparably well to unregularized ones on downstream tasks, indicating that the improvement in the agreement does not come at a cost for downstream performance.
When selecting regularized models, we choose ones with the strongest regularization scale hyperparameter that performs within $3$ $F_1$ points on the validation set compared to the unregularized model (cf.~details in \Cref{subsec:A_hs}). 
%

\section{The Cartography of Agreement}
\label{sec:agreement-cartography}

We have shown that by using a more appropriate correlation measure and applying regularization, the agreement of saliency methods increases significantly.
In this section, we are interested in finding out the cause of increased agreement obtained through applying regularization -- are there certain instance groups in the dataset that benefit the most, and if so, what changes in the representation space resulted in the increased agreement? 
We leverage methods from dataset cartography \cite{swayamdipta2020dataset} to distribute instances into \textit{easy-to-learn}, \textit{hard-to-learn}, and \textit{ambiguous} categories based on their prediction confidence and variability.
Concretely, if an instance exhibits \textbf{low} prediction variability and \textbf{high} prediction confidence between epochs, this implies that the model can quickly and accurately classify those instances, making them \textit{easy-to-learn}.
Instances that also exhibit \textbf{low} variability but \textbf{low} prediction confidence, align with the idea that the model is consistently unable to correctly classify them, making them \textit{hard-to-learn}.
Finally, instances that exhibit \textbf{high} variability and confidence close to the decision threshold indicate that the model is likely often changing its prediction between class labels for those instances, making them \textit{ambiguous}.
Since ambiguous instances are characterized by confidence near the prediction threshold, \citet{swayamdipta2020dataset} complement \textit{variability} and \textit{confidence} with another statistic introduced by \citet{chang2017active}, namely \textit{closeness}, defined as $c_i = p^{(i)} \cdot (1-p^{(i)})$, where $p^{(i)}$ is the average correct class probability of instance $x^{(i)}$ across all training epochs.
A \textbf{high} closeness value denotes that the instance is consistently near the decision boundary, and thus is a good indicator of ambiguity within the model. 

Intuitively, one would expect high agreement between saliency methods on instances that are easy to learn and low agreement otherwise.
However, when computing how agreement distributes across instance groups, we find the converse is true.
In unregularized models, we observe that easy-to-learn instances exhibit low average agreement, while ambiguous instances have a high average agreement.
In \Cref{tab:cart_dbert_full}, we report average agreement scores across all pairs of saliency methods on representative samples from each cartography group.\footnote{We select representative samples for each group through the relative frequency of their correct classification. If out of $5$ epochs, an instance was correctly classified $5$ times, it is representative of the \textit{easy-to-learn} category. If it was correctly classified $0$ times, it is representative of the \textit{hard-to-learn} category, and if the number of correct classifications is $2$ or $3$, it is representative of the \textit{ambiguous} category.}
We observe a clear distinction in agreement for both the base and regularized models, which is higher for ambiguous instances when compared to easy- and hard-to-learn instance groups.
Furthermore, we can observe a consistently high increase in agreement when the models are regularized across all instance groups for all datasets, indicating that regularization techniques reduce representation entanglement.

\begin{figure*}[t!]
\small
\centering
\begin{subfigure}{.24\textwidth}
  \centering
  \includegraphics[width=\linewidth]{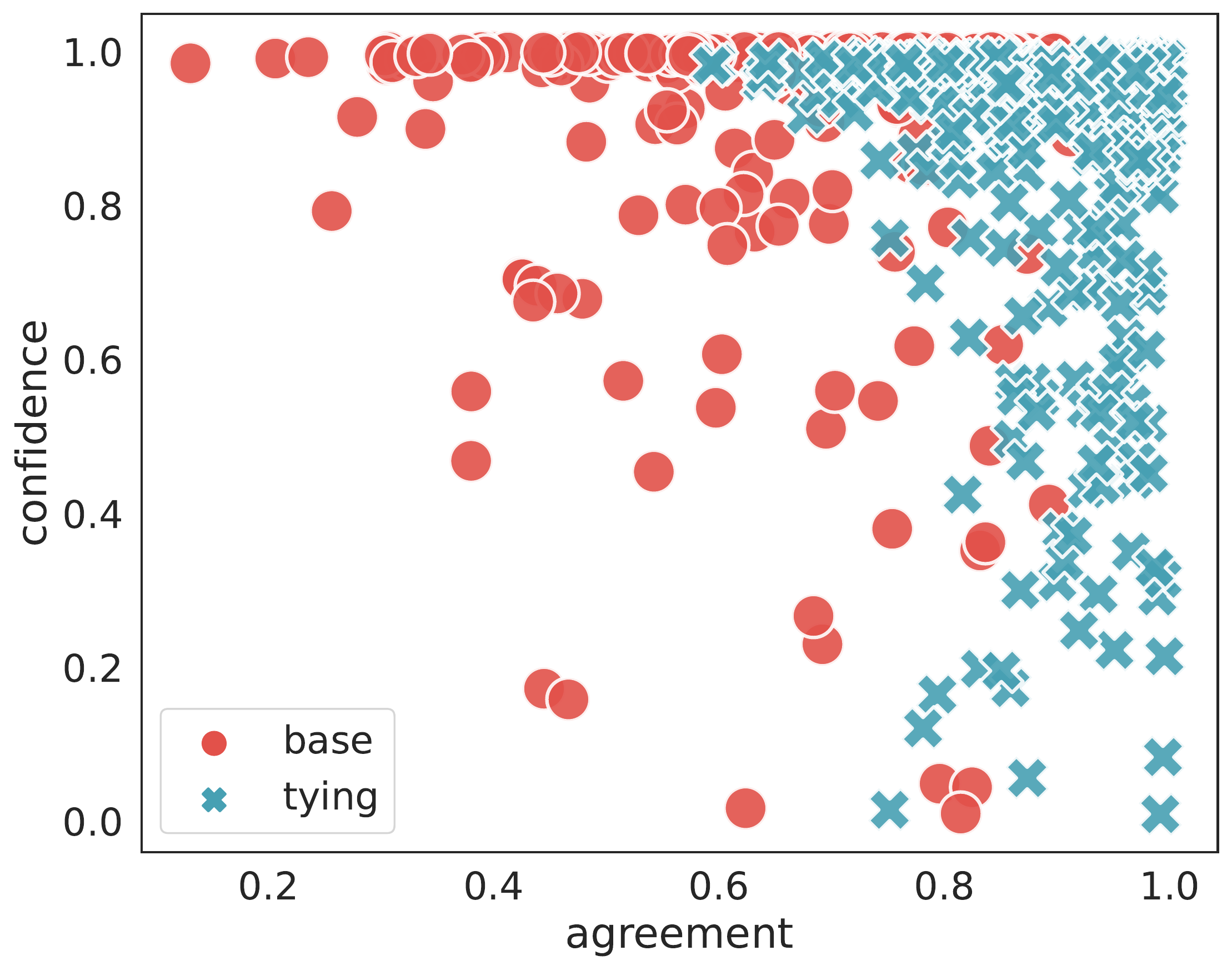}
  \caption{\textsc{jwa} -- SUBJ}
  \label{c}
\end{subfigure}
\begin{subfigure}{.24\textwidth}
  \centering
  \includegraphics[width=\linewidth]{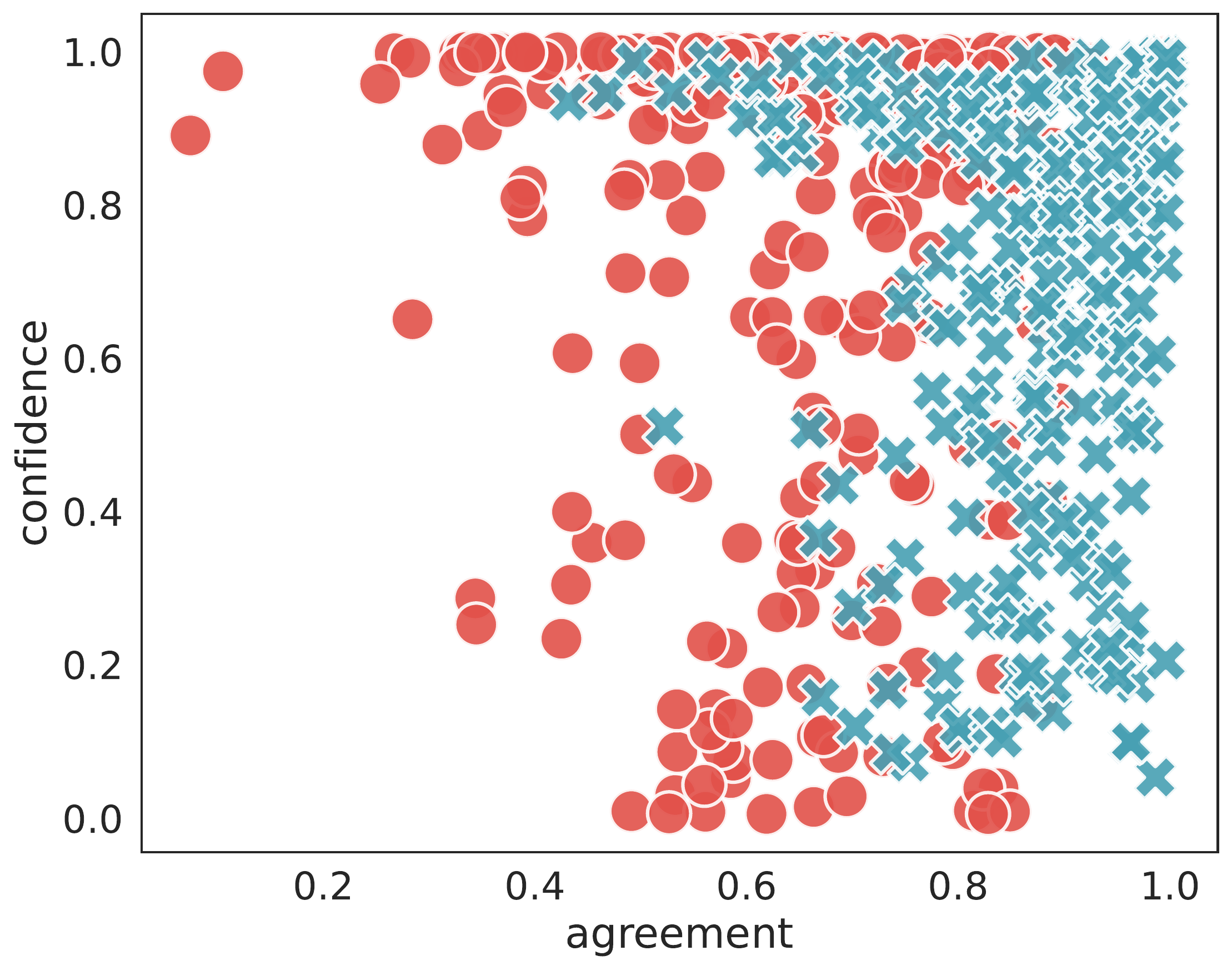}
  \caption{\textsc{jwa} -- SST}
  \label{fig:jwa-corr-sst}
\end{subfigure}
\begin{subfigure}{.24\textwidth}
  \centering
  \includegraphics[width=\linewidth]{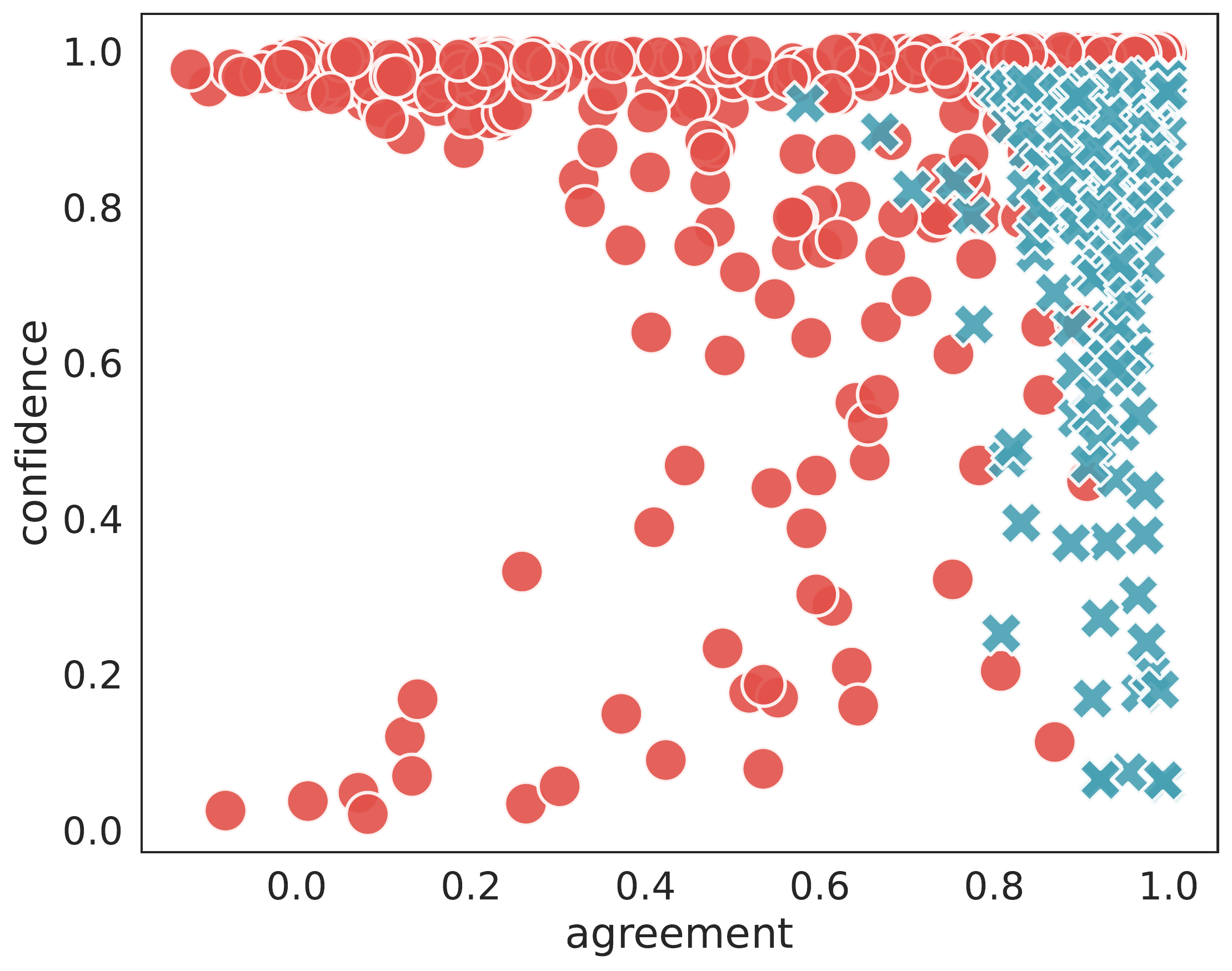}
  \caption{\textsc{jwa} -- TREC}
  \label{fig:jwa-corr-trec}
\end{subfigure}
\begin{subfigure}{.24\textwidth}
  \centering
  \includegraphics[width=\linewidth]{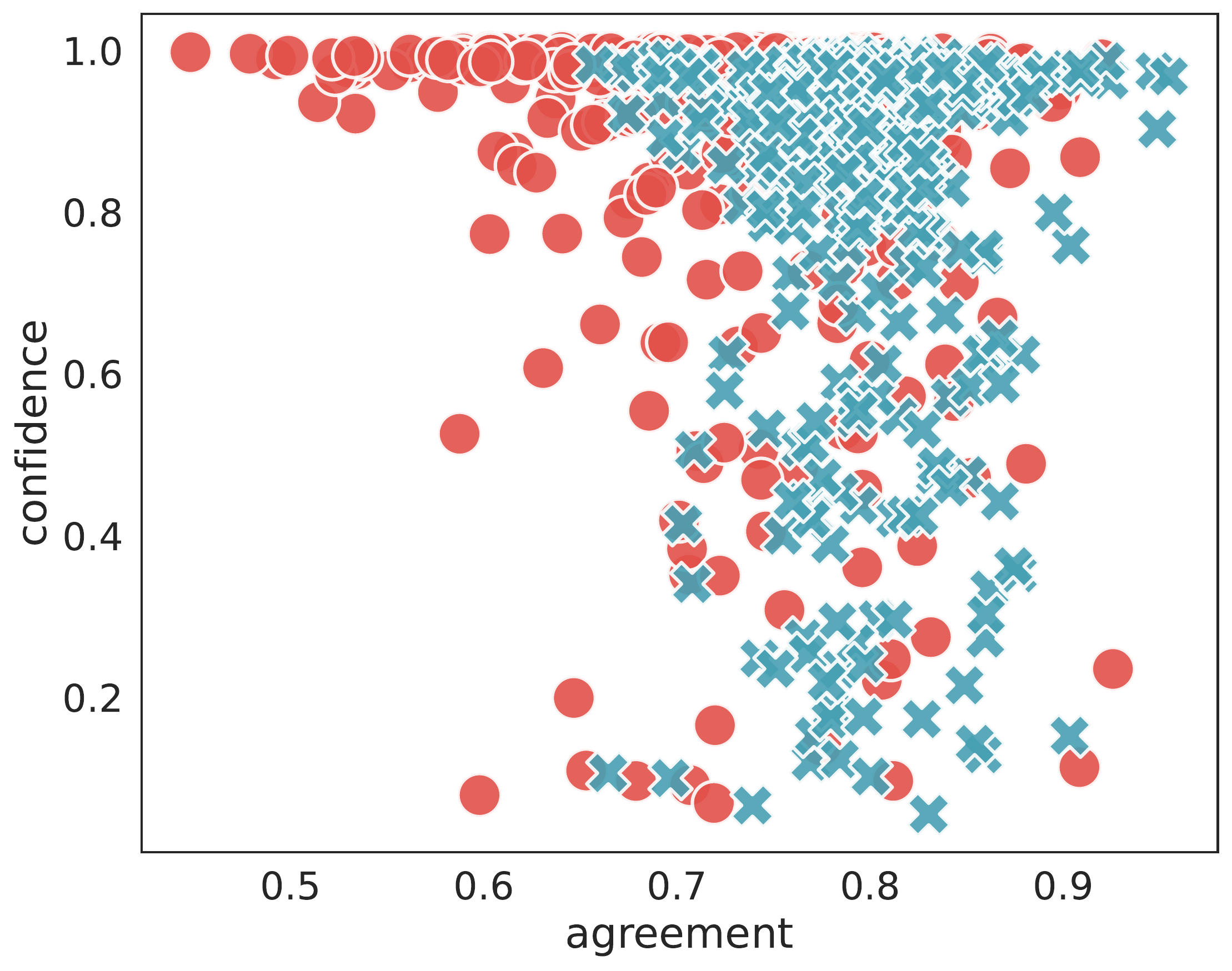}
  \caption{\textsc{jwa} -- IMDB}
  \label{fig:jwa-corr-imdb}
\end{subfigure}
\begin{subfigure}{.24\textwidth}
  \centering
  \includegraphics[width=\linewidth]{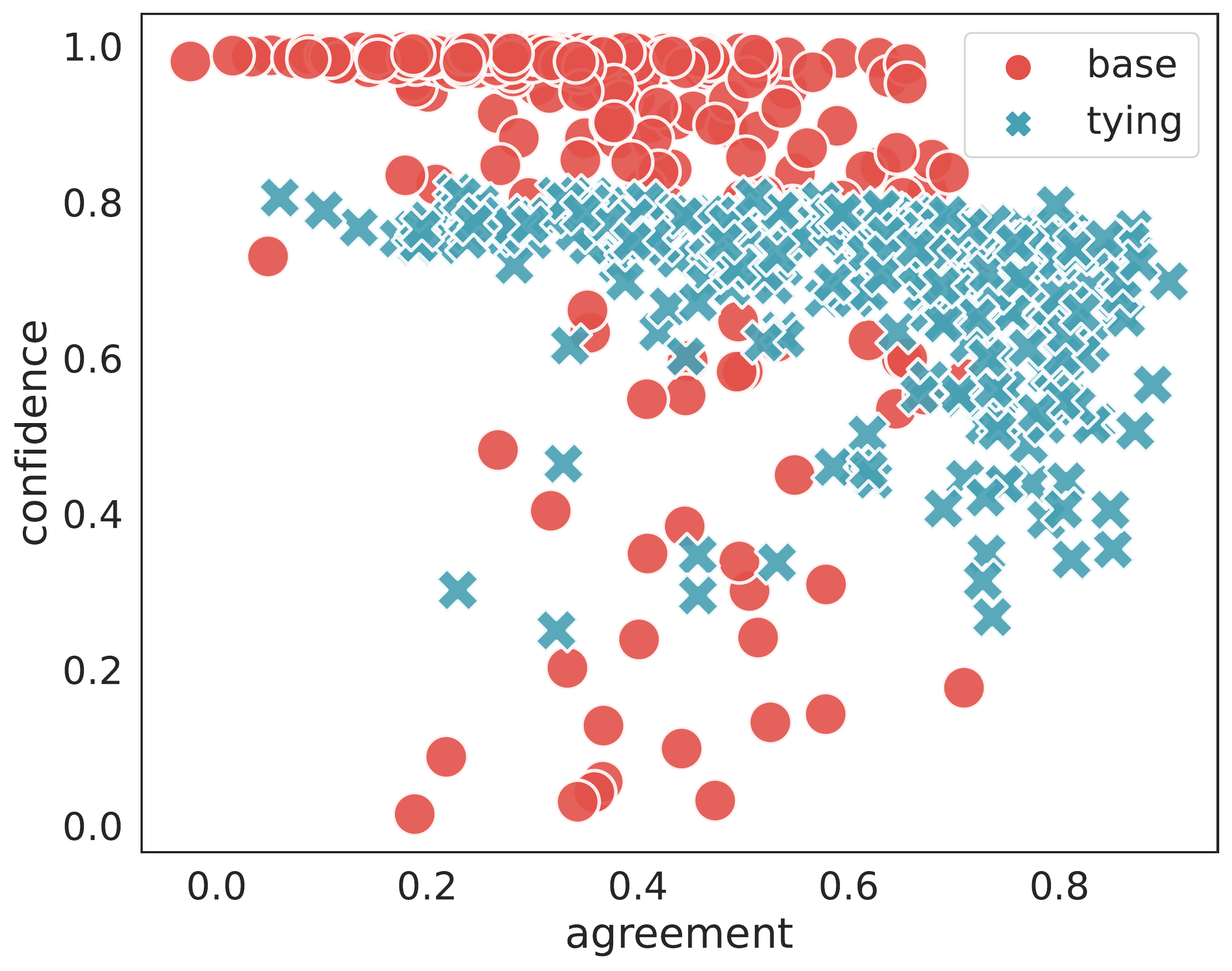}
  \caption{\textsc{dbert} -- SUBJ}
  \label{fig:dbert-corr-subj}
\end{subfigure}
\begin{subfigure}{.24\textwidth}
  \centering
  \includegraphics[width=\linewidth]{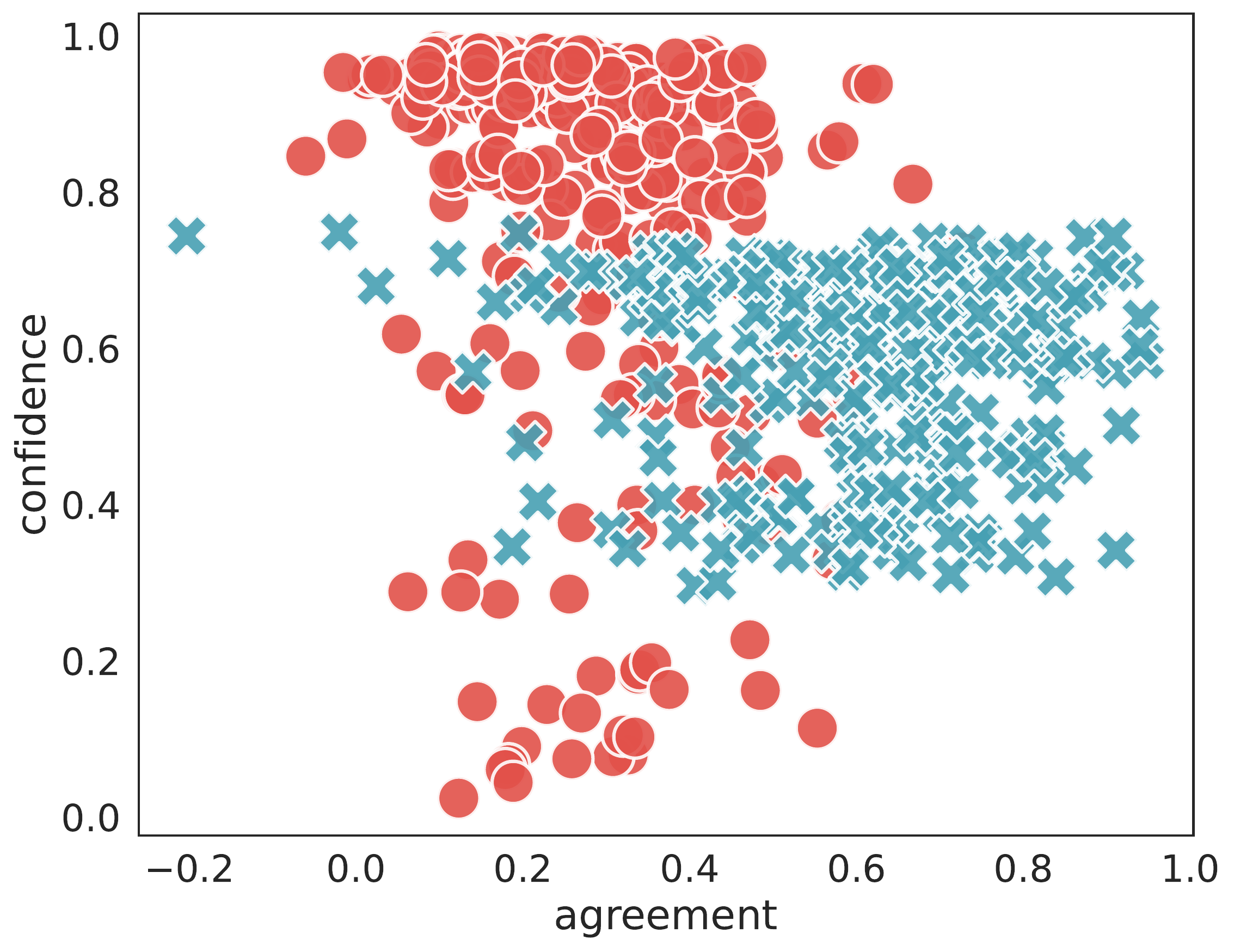}
  \caption{\textsc{dbert} -- SST}
  \label{fig:dbert-corr-sst}
\end{subfigure}
\begin{subfigure}{.24\textwidth}
  \centering
  \includegraphics[width=\linewidth]{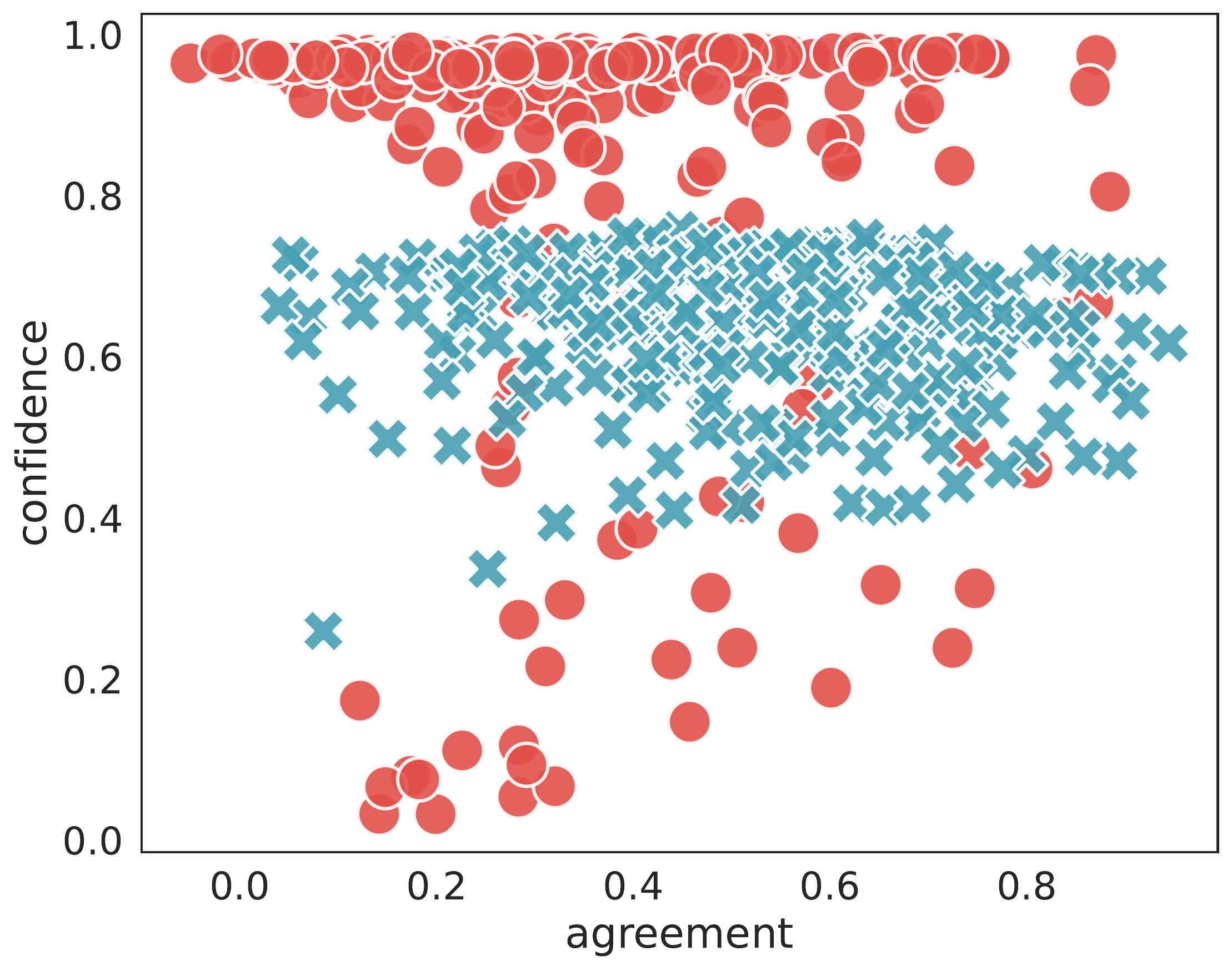}
  \caption{\textsc{dbert} -- TREC}
  \label{fig:dbert-corr-trec}
\end{subfigure}
\begin{subfigure}{.24\textwidth}
  \centering
  \includegraphics[width=\linewidth]{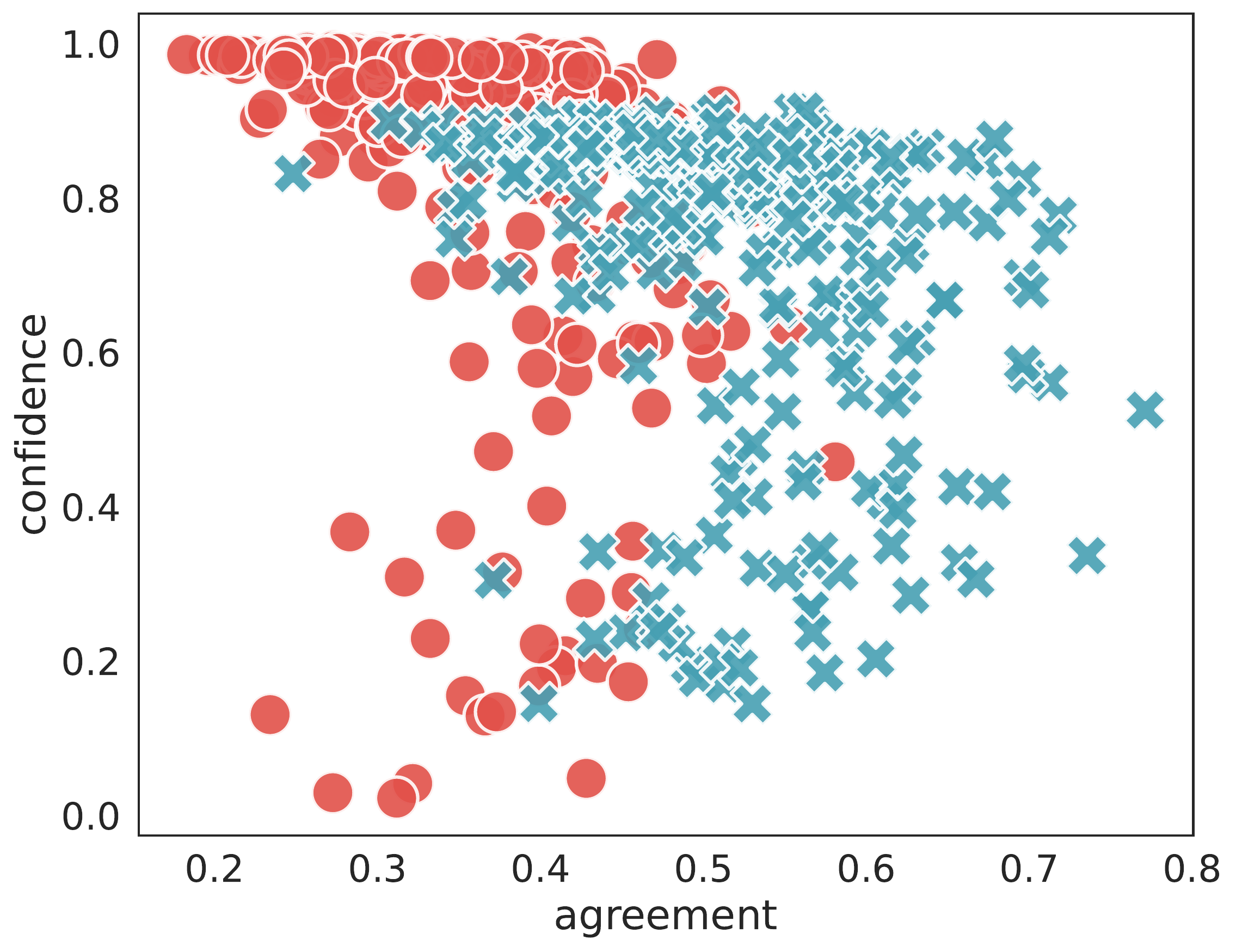}
  \caption{\textsc{dbert} -- IMDB}
  \label{fig:dbert-corr-imdb}
\end{subfigure}

\bigskip

\begin{subfigure}{.24\textwidth}
  \centering
  \includegraphics[width=\linewidth]{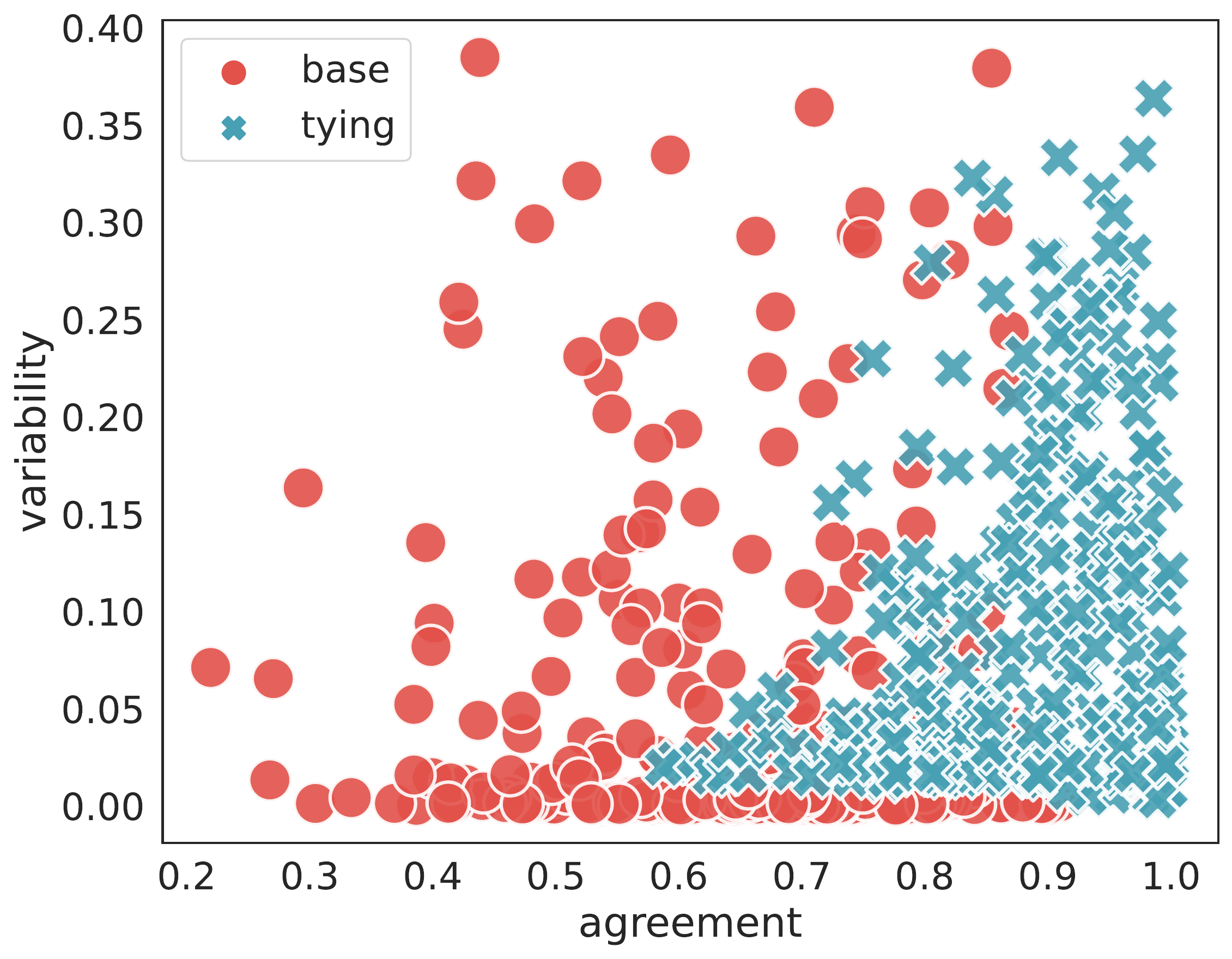}
  \caption{\textsc{jwa} -- SUBJ}
  \label{fig:jwa-var-subj}
\end{subfigure}
\begin{subfigure}{.24\textwidth}
  \centering
  \includegraphics[width=\linewidth]{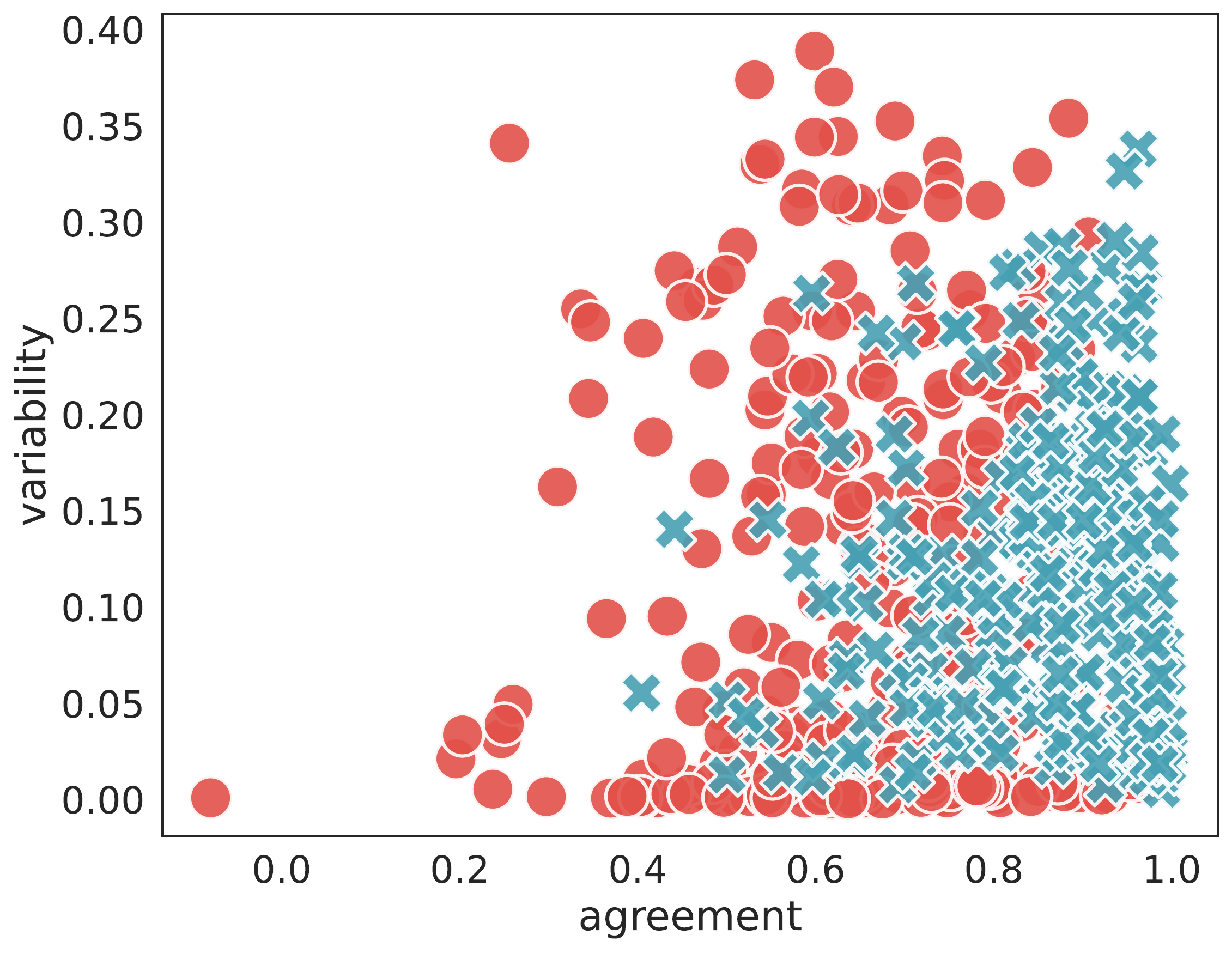}
  \caption{\textsc{jwa} -- SST}
  \label{fig:jwa-var-sst}
\end{subfigure}
\begin{subfigure}{.24\textwidth}
  \centering
  \includegraphics[width=\linewidth]{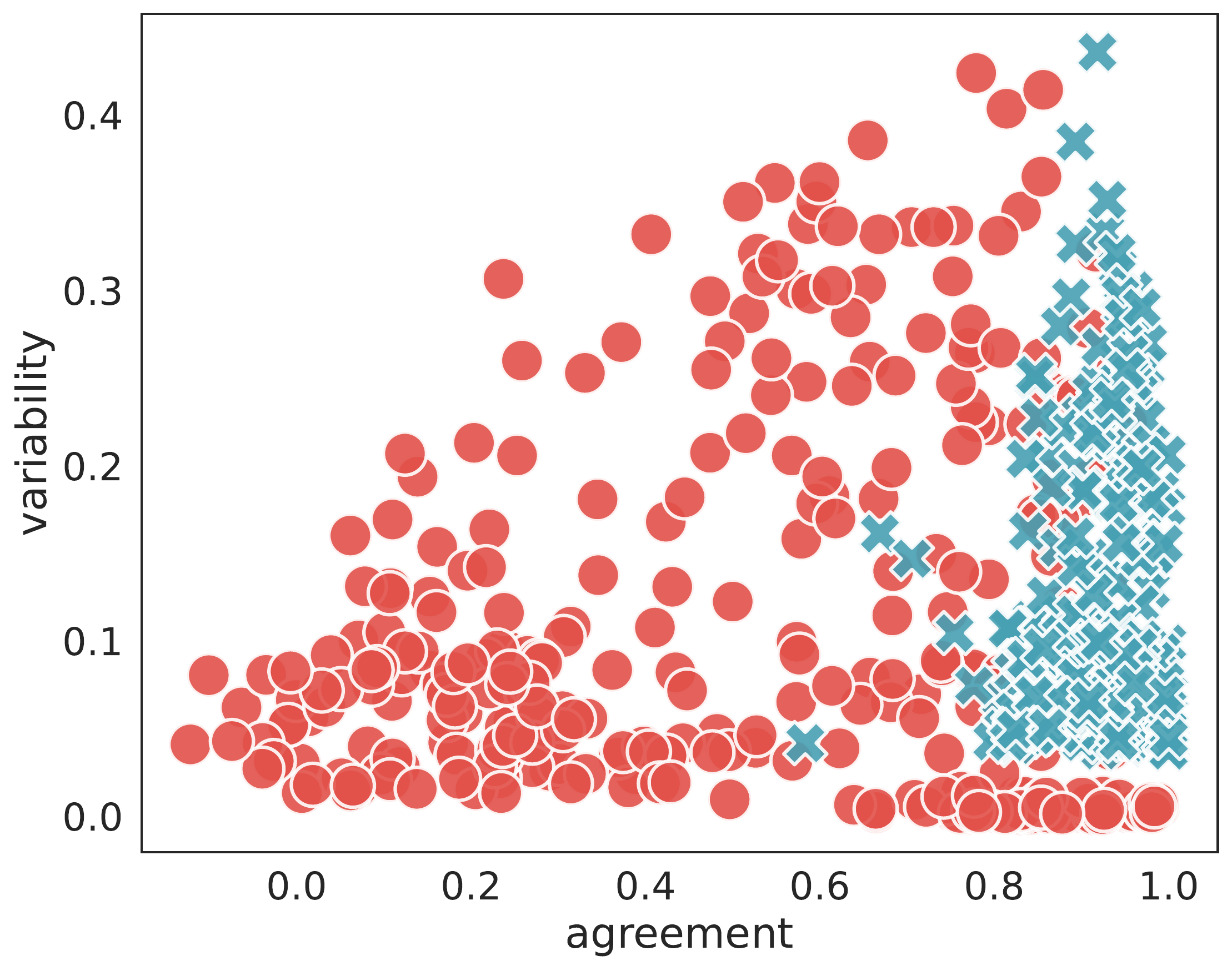}
  \caption{\textsc{jwa} -- TREC}
  \label{fig:jwa-var-trec}
\end{subfigure}
\begin{subfigure}{.24\textwidth}
  \centering
  \includegraphics[width=\linewidth]{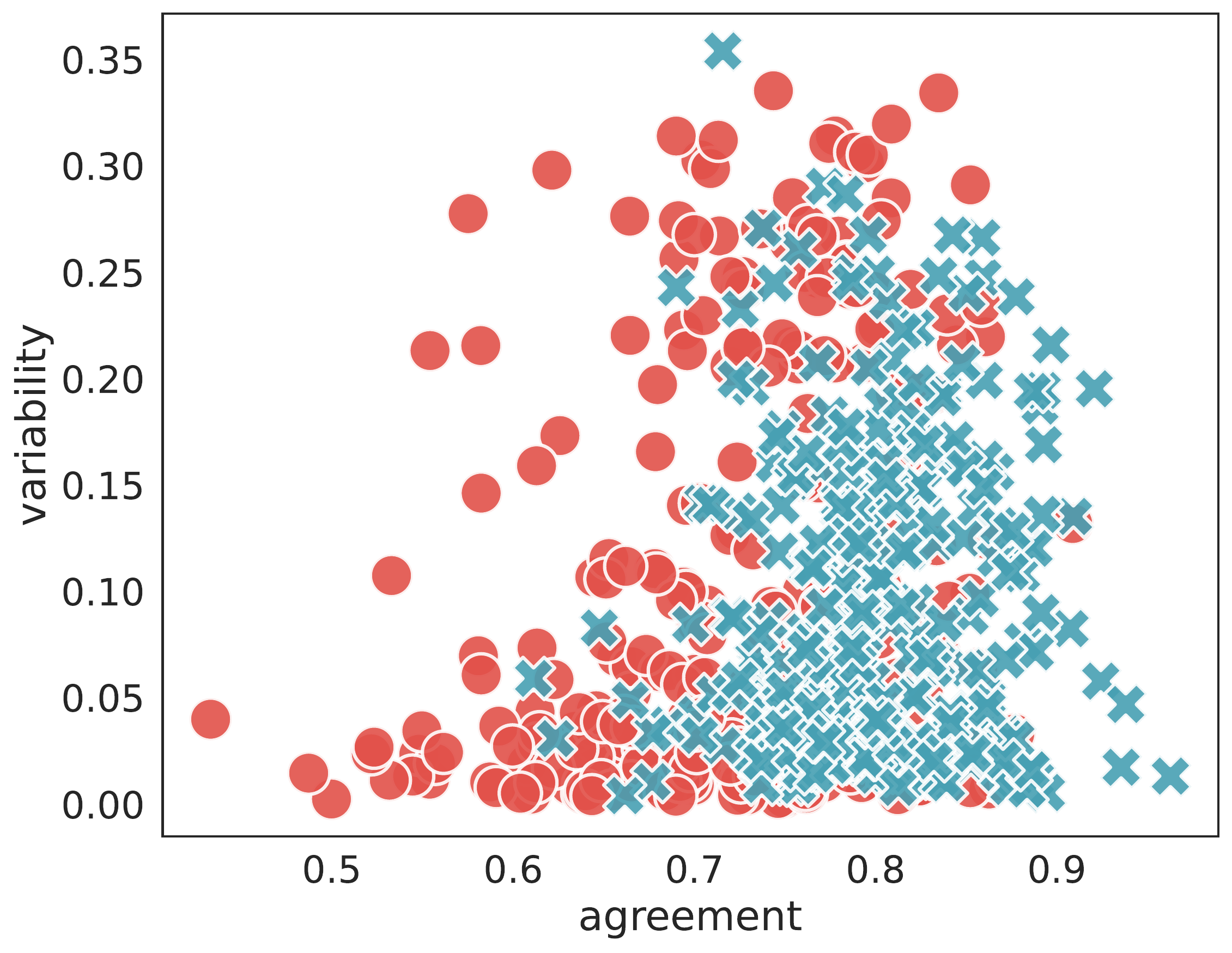}
  \caption{\textsc{jwa} -- IMDB}
  \label{fig:jwa-var-imdb}
\end{subfigure}
\begin{subfigure}{.24\textwidth}
  \centering
  \includegraphics[width=\linewidth]{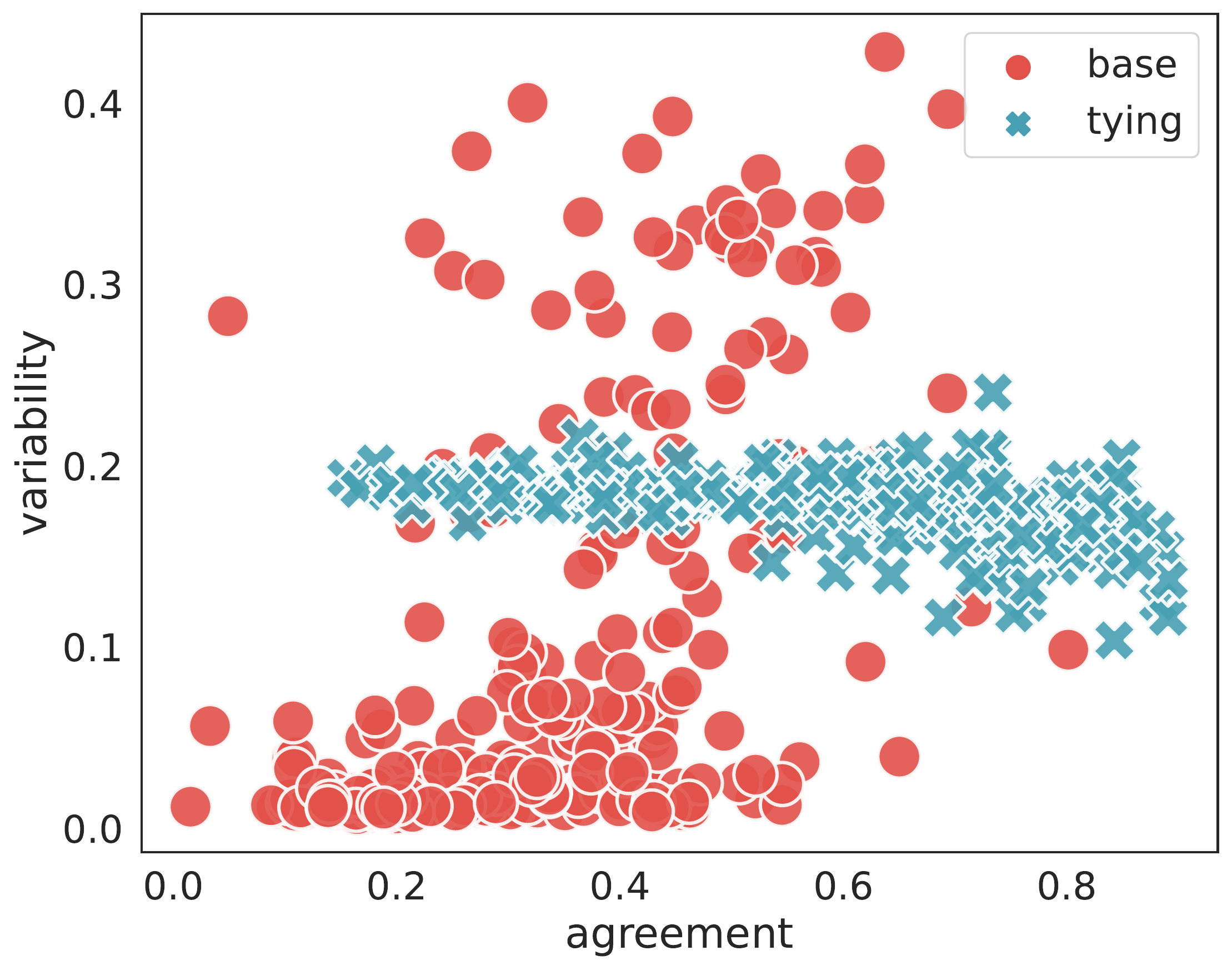}
  \caption{\textsc{dbert} -- SUBJ}
  \label{fig:dbert-var-subj}
\end{subfigure}
\begin{subfigure}{.24\textwidth}
  \centering
  \includegraphics[width=\linewidth]{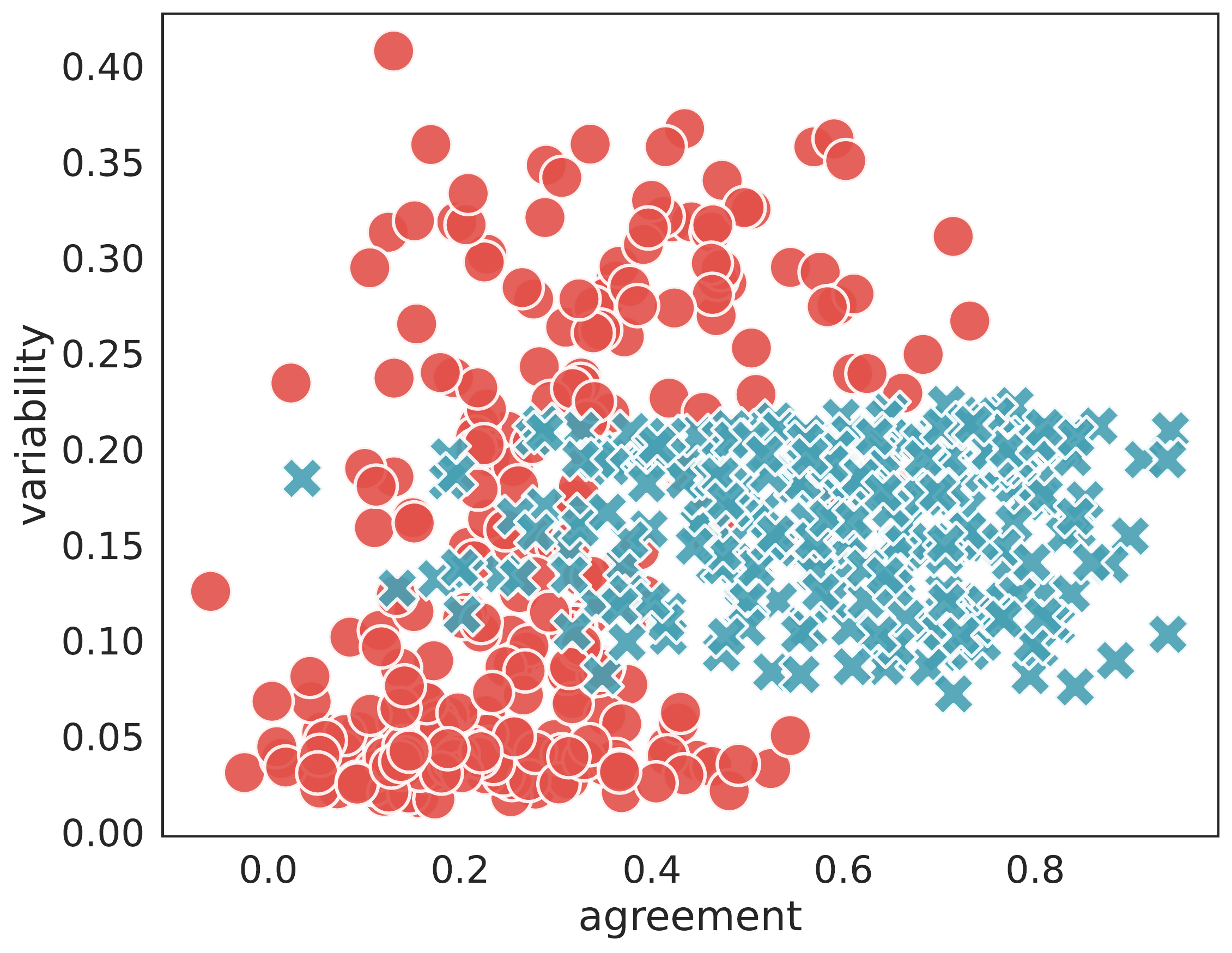}
  \caption{\textsc{dbert} -- SST}
  \label{fig:dbert-var-sst}
\end{subfigure}
\begin{subfigure}{.24\textwidth}
  \centering
  \includegraphics[width=\linewidth]{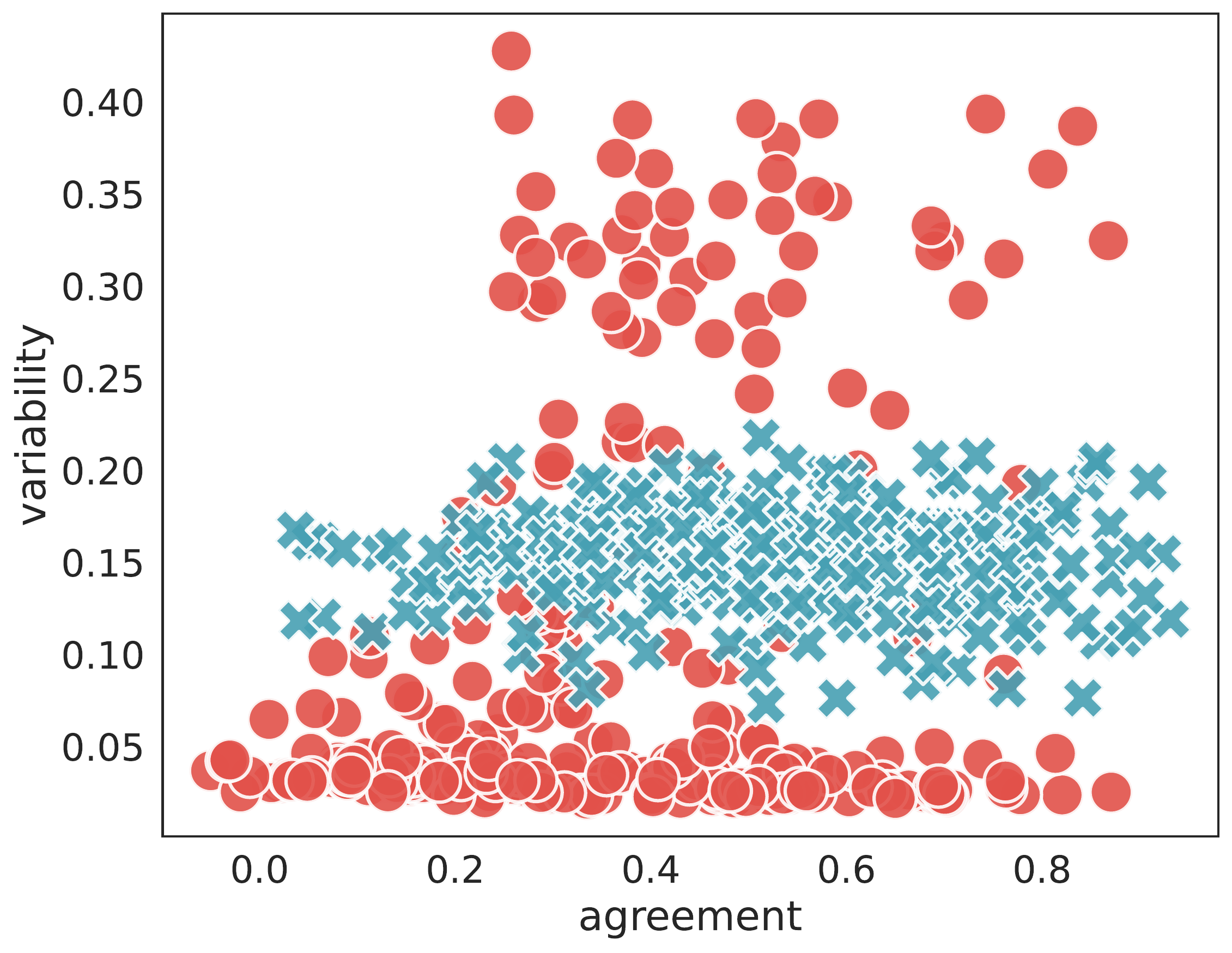}
  \caption{\textsc{dbert} -- TREC}
  \label{fig:dbert-var-trec}
\end{subfigure}
\begin{subfigure}{.24\textwidth}
  \centering
  \includegraphics[width=\linewidth]{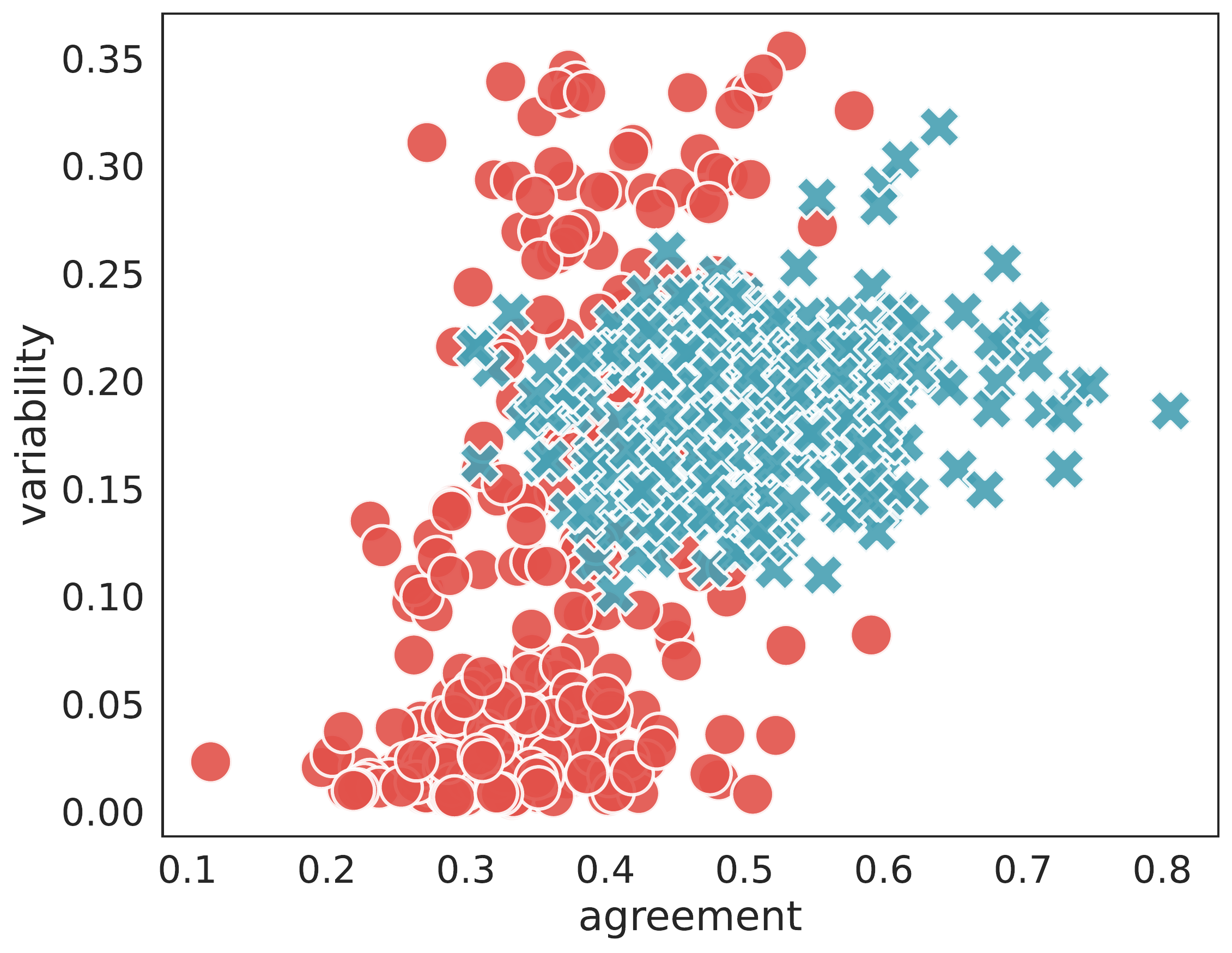}
  \caption{\textsc{dbert} -- IMDB}
  \label{fig:dbert-var-imdb}
\end{subfigure}
\caption{Relationship between agreement and cartography. Subfigures (a)--(h) and (i)--(p) show Pearson-$r$ agreement for each instance in the dataset with respect to \textit{confidence} and \textit{variability}, respectively. The red dots pertain to the unregularized model (\textsc{base}), and the blue crosses to the regularized model (\textsc{tying}).}
\label{fig:regularized-cartography}
\end{figure*}

One might wonder how the increase in agreement distributes across instances and dataset cartography attributes.
In \Cref{fig:regularized-cartography}, we visualize how the relationship between agreement and cartography attributes changes when the models are regularized.
We observe that for the \textsc{jwa} model, all datasets exhibit a consistent and significant increase in agreement.
Furthermore, we notice that for the \textsc{dbert} model, apart from increasing the agreement, regularization reduces the confidence of the model predictions and increases variability -- indicating that it reduces the known problem of overconfidence present in pre-trained language models.

\begin{table}
\small
\centering
\begin{tabular}{lrrrrrr}
\toprule
& \multicolumn{2}{c}{Easy} & \multicolumn{2}{c}{Amb} & \multicolumn{2}{c}{Hard} \\
\cmidrule(lr){2-3} \cmidrule(lr){4-5} \cmidrule(lr){6-7}
& B & T & B & T & B & T \\
\midrule
SUBJ & $.28$ & $\mathbf{.52}$\nospacetext{$^\dagger$} & $.48$ & $\mathbf{.74}$\nospacetext{$^\dagger$} & $.40$ & $\mathbf{.57}$\nospacetext{$^\dagger$} \\
SST & $.24$ & $\mathbf{.57}$\nospacetext{$^\dagger$} & $.36$ & $\mathbf{.63}$\nospacetext{$^\dagger$} & $.30$ & $\mathbf{.55}$\nospacetext{$^\dagger$} \\
TREC & $.33$ & $\mathbf{.49}$\nospacetext{$^\dagger$} & $.50$ & $\mathbf{.62}$\nospacetext{$^\dagger$} & $.32$ & $\mathbf{.40}$\nospacetext{$^\dagger$} \\
IMDB & $.34$ & $\mathbf{.48}$\nospacetext{$^\dagger$} & $.42$ & $\mathbf{.59}$\nospacetext{$^\dagger$} & $.36$ & $\mathbf{.51}$\nospacetext{$^\dagger$} \\
\bottomrule
\end{tabular}
\caption{Average agreement (Pearson-$r$) across saliency methods pairs per cartography groups. We select representative samples for each group based on the number of times a certain instance was classified correctly during training. We report average agreement for the unregularized model (B) and the one regularized by weight tying (T), with the numbers in \textbf{bold} indicating the higher agreement value among the two models. We averaged the results over $5$ runs. We ran one-sided Wilcoxon tests to check whether T is significantly better than B for a particular group. Significantly higher agreement values ($p < .05$) are marked with a $\dagger$.
}
\label{tab:cart_dbert_full}
\end{table}

\subsection{The Curvature of Agreement}
\label{sub:agr-curvature}
To better understand the cause of this distinction between various feature groups, we now analyze local curvature and density in the representation space.
We are interested in: (1) how densely the instances are distributed in the representation space across cartography categories and (2) whether the local space around an instance is sharp or smooth. 
For both models and all instances, we obtain sequence representations $h$ used as inputs to the decoder.
We estimate instance density as the average distance to the nearest instance in the dataset.
We estimate local smoothness around an instance representation as the $L_2$ norm of the gradient of the hidden representation with respect to the input embeddings. 
If the gradient norm is high, the local space is sharp and minor perturbations can have a large effect on the prediction probability.

In \Cref{tab:repr_dbert}, we report correlations between each of these two statistics and dataset cartography attributes.
We observe that for the unregularized model, there is a significant negative correlation between confidence and both gradient norm and minimum distance to the nearest example, indicating that the local space around easy instances is smooth and densely populated.
On the other hand, there is a high positive correlation between both closeness and variability and both gradient norm and minimum distance to the nearest example -- indicating that the local space around ambiguous instances is sharp and sparsely populated.
When we turn our attention to the regularized model, we observe that the correlation between the gradient norm and any of the cartography attributes vanishes, while the correlations between distance and the attributes are reduced in absolute value and their sign is flipped.

From these observations, we hypothesize that the cause of low agreement on easy-to-learn instances is the multitude of possible explanations as to why such an instance should be correctly classified.
From the viewpoint of \textit{plausibility}, this hypothesis is in line with the Rashomon effect \cite{breiman2001statistical} which is about there often existing a multitude of adequate descriptions that end up with the same error rate, or in our case, prediction probability -- however it should not apply to \textit{faithfulness}, as a single model instance should adhere to a single explanation.
However, due to a plethora of corroborating evidence for easy-to-learn instances, the representation space
around them is smooth to such an extent that perturbations do not significantly affect the prediction probability, which in turn adversely affects gradient- and propagation-based explanation methods.
The converse is true for ambiguous instances, where we hypothesize the model observes evidence for both classes and is unable to reach a confident decision.
However, this difficulty in reaching a decision also causes saliency methods to have a precise definition of what the evidence is -- as the local curvature is sharp, and any minor perturbation could significantly affect prediction probability.
We believe that local curvature statistics could be used as a metric for measuring whether a trained model is better suited to analysis through explainability methods.


\begin{table}
\small
\centering
\begin{tabular}{lrrrrrr}
\toprule
& \multicolumn{2}{c}{Conf} & \multicolumn{2}{c}{Close} & \multicolumn{2}{c}{Var} \\
\cmidrule(lr){2-3} \cmidrule(lr){4-5} \cmidrule(lr){6-7}
& B & T & B & T & B & T \\
\midrule
Grad norm & $-.39$ & $-.02$ & $.52$ & $.05$ & $.46$ & $.06$ \\
Min dist & $-.53$ & $.25$ & $.72$ & $-.39$ & $.64$ & $.16$ \\
\bottomrule
\end{tabular}
\caption{Correlations (Pearson-$r$) between local curvature statistics in the representation space and cartography attributes. We report average gradient norms (\textit{grad norm}) of the hidden representation with respect to the input embeddings and average distance to the nearest instance (\textit{min dist}). Columns correspond to the cartography attributes: \textit{conf} -- confidence, \textit{close} -- closeness, and \textit{var} -- confidence variance. We report the results for the \textsc{base} model (B) and the model regularized by \textsc{tying} (T). Results reported are averages over all datasets.}
\label{tab:repr_dbert}
\end{table}

\section{Conclusion}
\label{sec:conclusion}

We analyzed two prototypical models from different families in \textsc{jwa} and \textsc{dbert} with the goal of finding out the cause of low agreement between saliency method interpretations.
We first took a closer look at the previously used rank-order correlation metric and demonstrated that it is prone to exhibiting a high difference in agreement for small changes in importance scores. As an alternative, we used a linear correlation metric that is robust to small importance perturbations and demonstrated that it exhibits consistently higher agreement scores.
Taking a step further, we applied two regularization techniques, \textsc{tying} and \textsc{conicity}, originally aimed at increasing faithfulness of attention explanations, with the hypothesis that the issue underpinning disagreements and unfaithfulness is the same -- representation entanglement in the hidden space.
We showed that regularization consistently and significantly improves agreement scores across all models and datasets with a minimal penalty for classification performance.
Having demonstrated that it is possible to improve upon the low agreement scores, we attempted to offer intuition on which instance categories saliency methods agree the least and show that surprisingly, \textit{easy-to-learn} instances are \textit{hard to agree} on.
Lastly, we offered insights into how the representation space morphs when regularization is applied and linked these findings with dataset cartography categories, paving the way for further work on understanding what properties of neural models affect interpretability.

\section*{Limitations}
\label{sec:limitations}

Our work has a number of important limitations that affect the conclusions we can draw from it.
First and foremost, evaluating the faithfulness of model interpretations is problematic as we do not have ground truth annotations for token importances.
Thus, when applying the \textit{agreement-as-evaluation} paradigm, we implicitly assume that most saliency methods are close to the truth -- an assumption that we cannot verify.
However, every method of evaluating faithfulness has its own downsides. Token and representation erasure runs the risk of drawing conclusions from corrupted inputs that fall off the data manifold.
We argue that while \textit{agreement-as-evaluation} is far from an ideal way of evaluating faithfulness, it still increases credibility when used along with other techniques.

Secondly, our work is limited both with respect to the datasets and models considered.
Specifically, we only evaluate one Transformer-based model from the masked language modeling family, and it is entirely possible that the findings do not generalize to models pre-trained on different tasks.
Also, we only consider single sequence classification datasets -- mainly due to the fact that the issues with the faithfulness of attention were most prevalent in those setups, which we assumed would be the same for agreement due to the same hypothesized underlying issue.
We believe that tasks that require retention of token-level information in hidden states, such as sequence labeling and machine translation, would exhibit higher agreement overall, even without intervention through regularization. We leave this analysis for future work.




\bibliography{interpretability_agreement}
\bibliographystyle{acl_natbib}

\clearpage
\appendix

\section{Reproducibility}
\label{sec:A}

\subsection{Experimental results}

\subsubsection{Setup}
For both \textsc{jwa} and \textsc{dbert}, we use the same preprocessing pipeline on all four datasets. First, we filter out instances with fewer than three tokens to achieve stable agreement evaluation.\footnote{If a sequence consists of only two tokens, rank-correlation with Kendall-$\tau$ will either result in a perfect match, or completely different observations as swapping the two ranks leads to an inverse ranking.}
Next, we lowercase the tokens, remove non-alphanumeric tokens, and truncate the sequence to $200$ tokens if the sequence length exceeds this threshold. We set the maximum vocabulary size to $20$k for models which do not leverage subword vocabularies.

\subsubsection{Validation set performance}
We report the validation set performance in \Cref{tab:f1_scores_val}.

\subsubsection{Computing infrastructure}
We conducted our experiments on $2 \times$ \textit{AMD Ryzen Threadripper 3970X 32-Core Processors} and $2 \times$ \textit{NVIDIA GeForce RTX 3090} GPUs with $24$GB of RAM. We used \textit{PyTorch} version $1.9.0$ and CUDA $11.4$.

\subsubsection{Average runtime}
\Cref{tab:runtime} shows the average experiment runtime for each model across the datasets we used.

\begin{table}[t!]
\small
\centering
\begin{tabular}{lrrrr}
\toprule
& & Base & Conicity & Tying \\
\midrule
\multirow{4}{*}{\rotatebox[origin=c]{90}{\textsc{dbert}}}
& SUBJ & $.94$ & $.89$ & $.94$ \\
& SST & $.85$ & $.85$ & $.85$ \\
& TREC & $.94$ & $.89$ & $.91$ \\
& IMDB & $.90$ & $.89$ & $.89$ \\
\midrule
\multirow{4}{*}{\rotatebox[origin=c]{90}{\textsc{jwa}}}
& SUBJ & $.93$ & $.90$ & $.91$ \\
& SST & $.82$ & $.79$ & $.81$ \\
& TREC & $.91$ & $.87$ & $.89$ \\
& IMDB & $.90$ & $.87$ & $.88$ \\
\bottomrule
\end{tabular}
\caption{$F_1$ scores on \textit{validation} sets across datasets for \textsc{dbert} and \textsc{jwa}. We average the results over $5$ runs with different seeds. The scores pertain to the same experiments as in \Cref{tab:f1_scores}, where we report test $F_1$ scores.}
\label{tab:f1_scores_val}
\end{table}

\begin{table}
\centering
\small
\begin{tabular}{lrr}
\toprule
& \textsc{jwa} & \textsc{dbert} \\
\midrule
SUBJ & $3.4$ & $11.2$ \\
SST & $2.7$ & $8.9$ \\
TREC & $1.2$ & $3.7$  \\
IMDB & $6.1$ & $107.5$ \\
\end{tabular}
\caption{Experiment duration in minutes for both models across datasets. We report the average runtime over $5$ different runs.}
\label{tab:runtime}
\end{table}

\subsubsection{Number of parameters}
The \textsc{jwa} and \textsc{dbert} models that we used contained $1,714,951$ and $66,954,241$ trainable parameters, respectively.

\subsection{Hyperparameter search}
\label{subsec:A_hs}
We used the following parameter grids for \textsc{jwa}: $[10^{-1}, 10^{-2}, 10^{-3}, 10^{-4}, 10^{-5}, 10^{-6}]$ for learning rate,  and $[50, 100, 150, 200]$ for the hidden state dimension. We yield best average results on validation sets across all datasets when the learning rate is set to $10^{-3}$ and the hidden size is set to $150$. For \textsc{dbert}, we find that the most robust initial learning rate on the four datasets is $2 \times 10^{-5}$, among the options we explored $[5 \times 10^{-4}, 10^{-4}, 10^{-5}, 2 \times 10^{-5}, 5 \times 10^{-5}, 10^{-6}]$. Additionally, we clip the gradients for both models such that the gradient norm $\le 1$. We use the Adam \cite{kingma-ba-2015-adam} optimizer for \textsc{jwa} and AdamW \cite{loshchilov-hutter-2017-fixing} for \textsc{dbert}. We run both models for $5$ epochs and repeat the experiments $5$ times with different seeds: $[1, 2, 3, 4, 5]$.

For regularization methods, we conducted a grid search with parameter grid $[0.1, 0.3, 0.5, 1, 5, 10]$ for \textsc{conicity} and $[0.1, 0.3, 0.5, 1, 5, 10, 20]$ for \textsc{tying}. We select the models with the strongest regularization scale, which is within $3$ $F_1$ points from the unregularized model. \Cref{tab:reg_params} shows the selected values for each model across all datasets.

\begin{table}
\small
\centering
\begin{tabular}{lrrrrrr}
\toprule
& \multicolumn{2}{c}{\textsc{jwa}} & \multicolumn{2}{c}{\textsc{dbert}} \\
\cmidrule(lr){2-3} \cmidrule(lr){4-5}
& C & T & C & T \\
\midrule
SUBJ & $1.$ & $1.$ & $5.$ & $1.$ \\
SST & $1.$ & $0.5$ & $0.1$ & $0.5$ \\
TREC & $1.$ & $1.$ & $0.1$ & $0.3$  \\
IMDB & $0.3$ & $1.$ & $1.$ & $1.$ \\
\bottomrule
\end{tabular}
\caption{Selected hyperparameter values for \textsc{conicity} (C) and \textsc{tying} (T).}
\label{tab:reg_params}
\end{table}

\begin{table}
\centering
\small
\begin{tabular}{lrr}
\toprule
& \textsc{jwa} & \textsc{dbert} \\
\midrule
SUBJ & $3.4$ & $11.2$ \\
SST & $2.7$ & $8.9$ \\
TREC & $1.2$ & $3.7$  \\
IMDB & $6.1$ & $107.5$ \\
\end{tabular}
\caption{Experiment duration in minutes for both models across datasets. We report the average runtime over $5$ different runs.}
\label{tab:runtime}
\end{table}

\subsection{Dataset statistics}
We report the number of instances per split for each dataset in \Cref{tab:dataset_stats}. We note that all of the datasets we used contain predominantly texts in English.

\begin{table}[t!]
\small
\centering
\begin{tabular}{lrrrr}
\toprule
& Train & Validation & Test & Total \\
\midrule
SUBJ & $7,000$ & $1,000$ & $2,000$ & $10,000$\\
SST & $6,819$ & $868$ & $1,810$ & $9,497$ \\
TREC & $1,987$ & $159$ & $486$ & $2,632$ \\
IMDB & $17,212$ & $4,304$ & $4,363$ & $25,879$ \\
\end{tabular}
\caption{Number of instances in each split and the total number of instances in each dataset after we excluded too short examples (see \cref{subsec:datasets}).}
\label{tab:dataset_stats}
\end{table}

\section{Additional experiments}
\label{sec:B}

\begin{figure}[h]

\centering
\includegraphics[width=\linewidth]{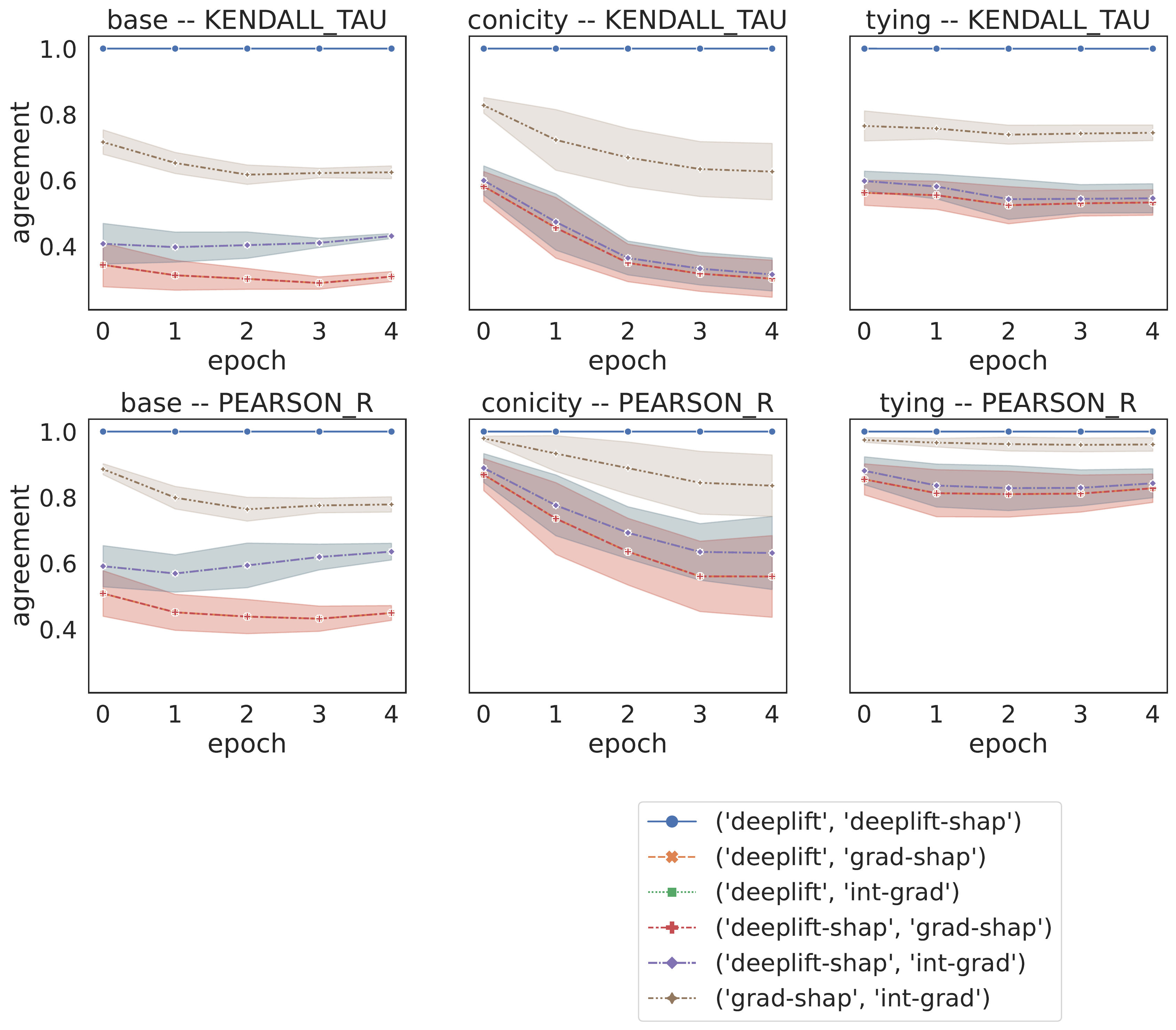}
\caption{\textsc{jwa} -- SUBJ}
\label{fig:corr-jwa-subj}
\end{figure}

\begin{figure}[h]
\centering
\includegraphics[width=\linewidth]{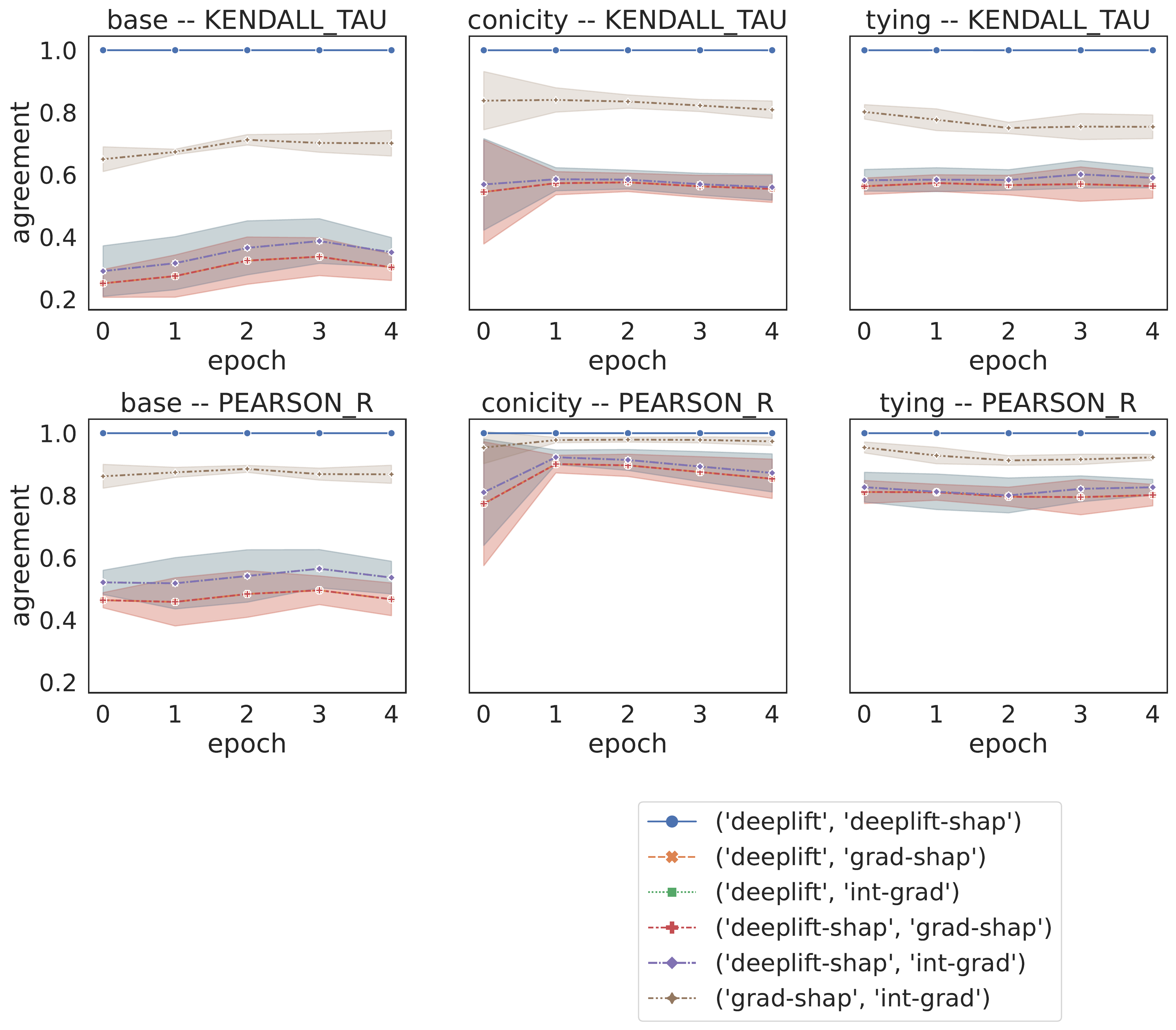}
\caption{\textsc{jwa} -- SST}
\label{fig:corr-jwa-sst}
\end{figure}

\begin{figure}
\centering
\includegraphics[width=\linewidth]{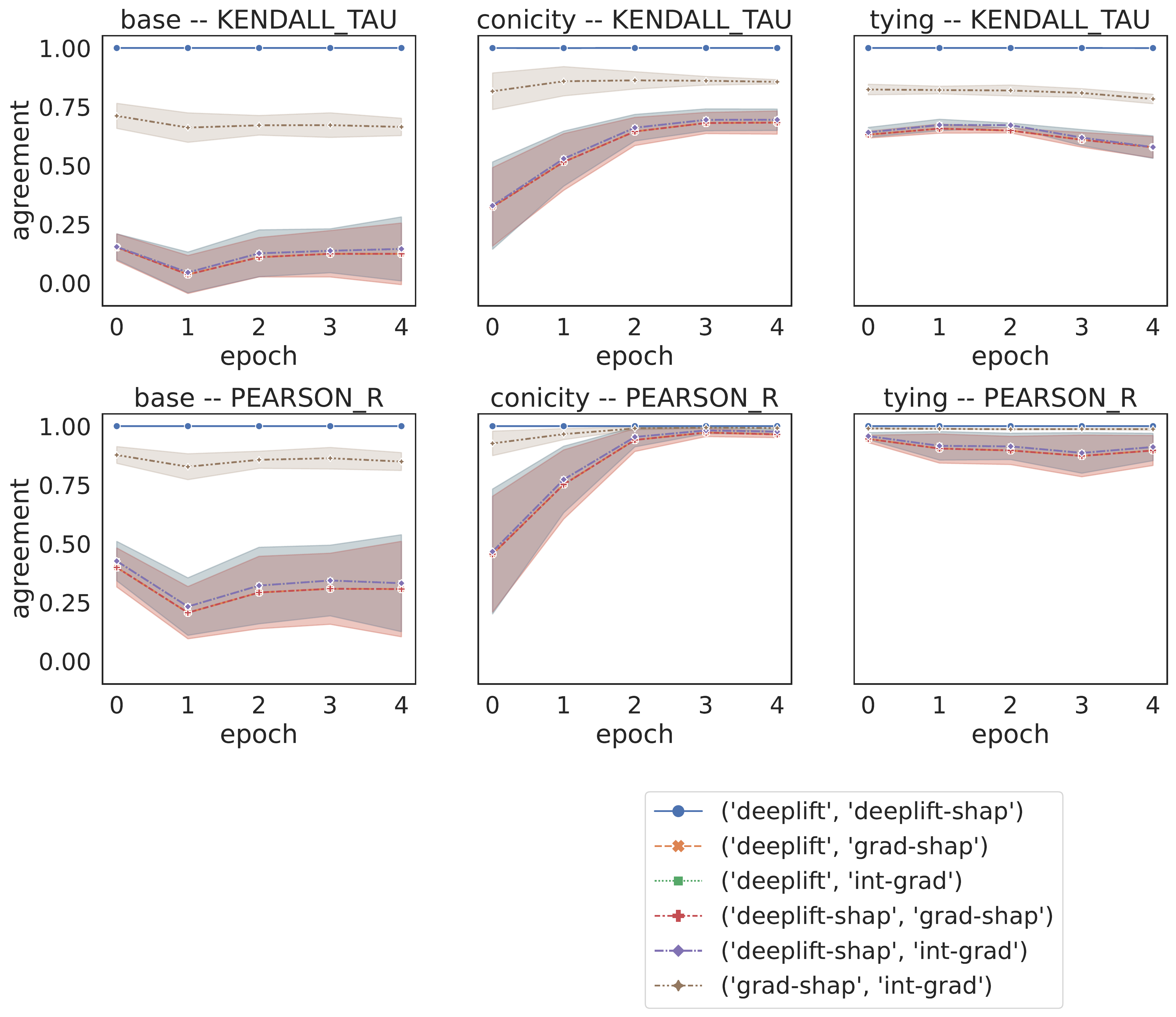}
\caption{\textsc{jwa} -- TREC}
\label{fig:corr-jwa-trec}
\end{figure}

\begin{figure}
\centering
\includegraphics[width=\linewidth]{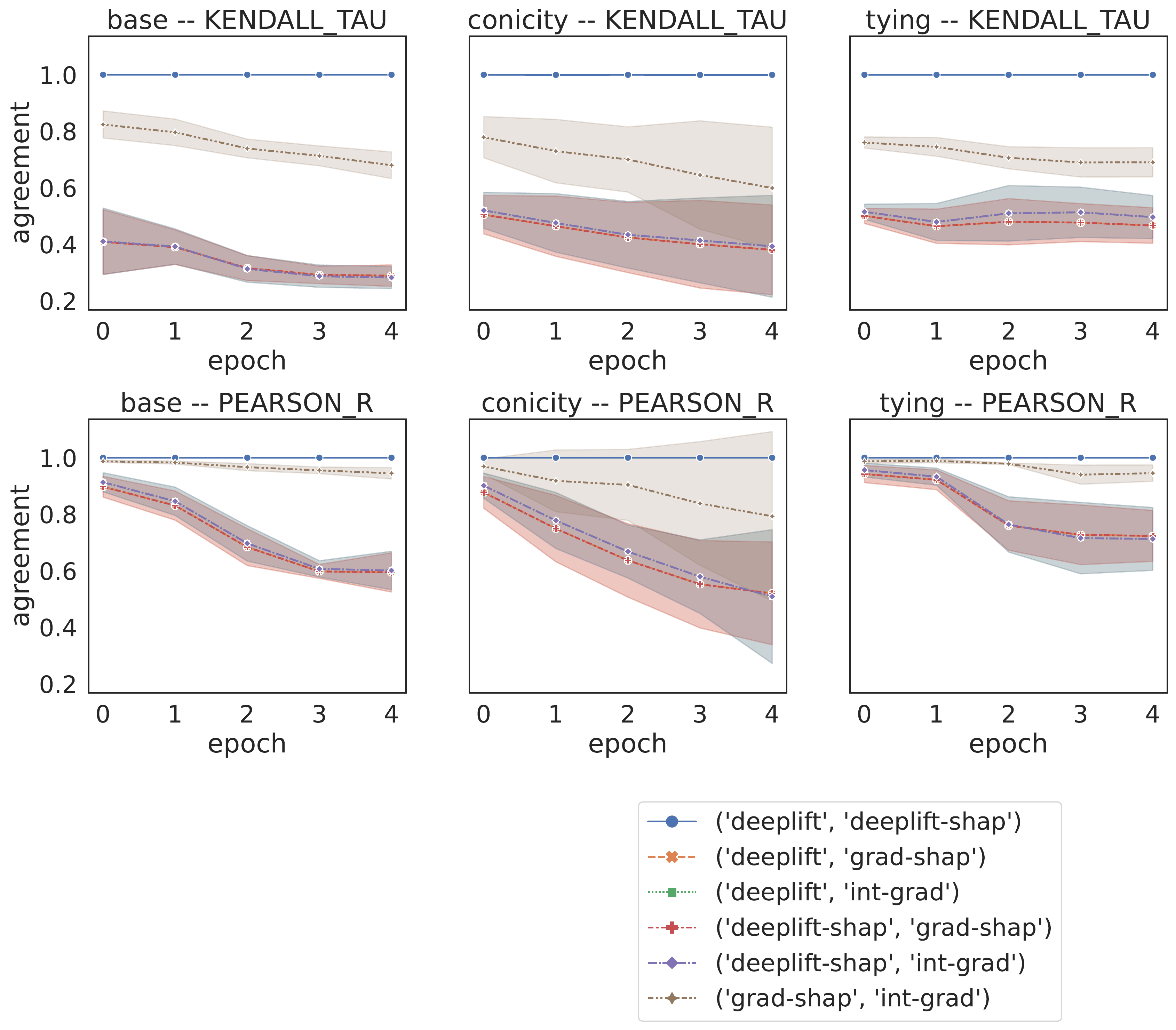}
\caption{\textsc{jwa} -- IMDB}
\label{fig:corr-jwa-imdb}
\end{figure}

\begin{figure}
\centering
\includegraphics[width=\linewidth]{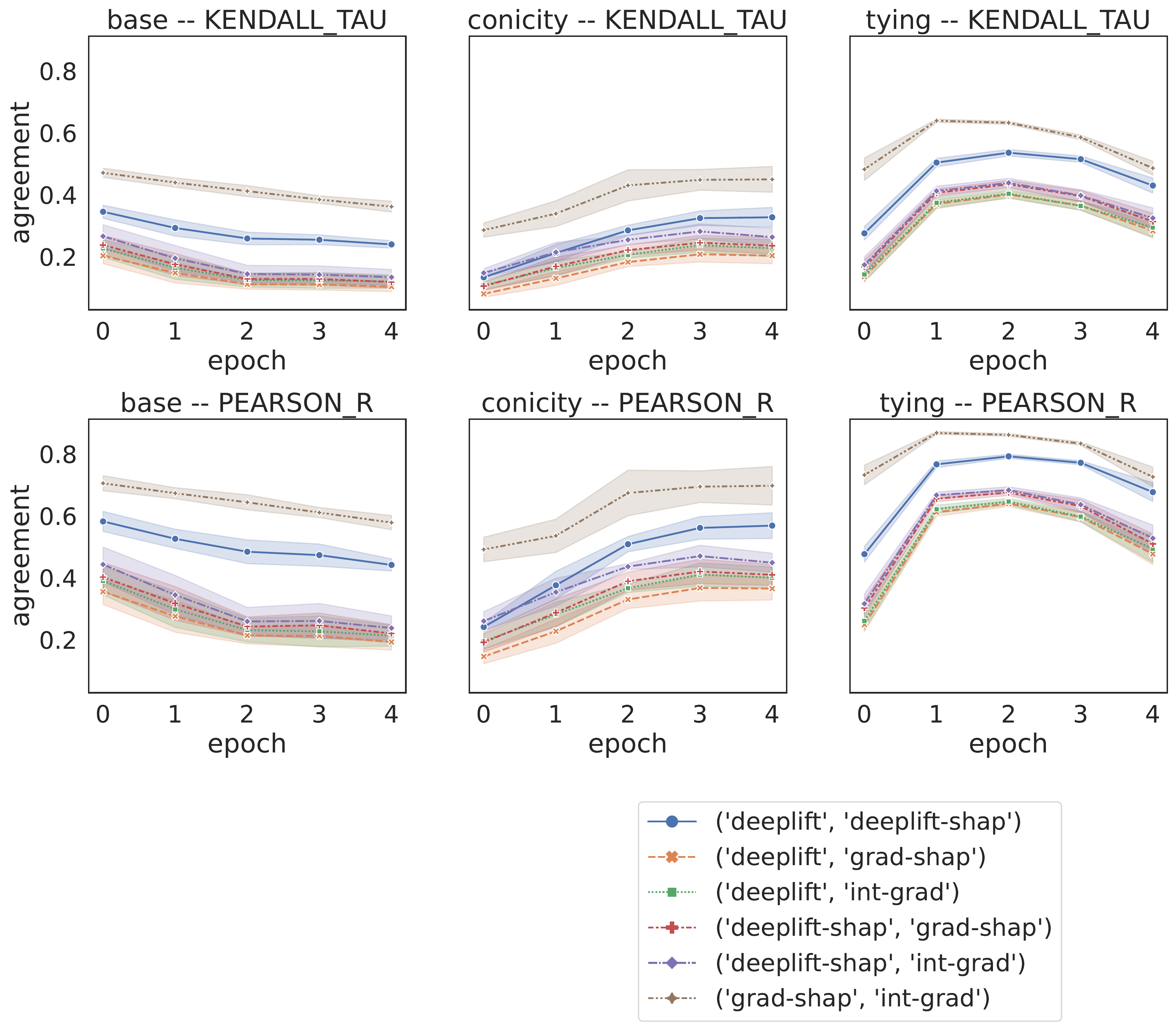}
\caption{\textsc{dbert} -- SUBJ}
\label{fig:corr-dbert-sst}
\end{figure}

\begin{figure}
\centering
\includegraphics[width=\linewidth]{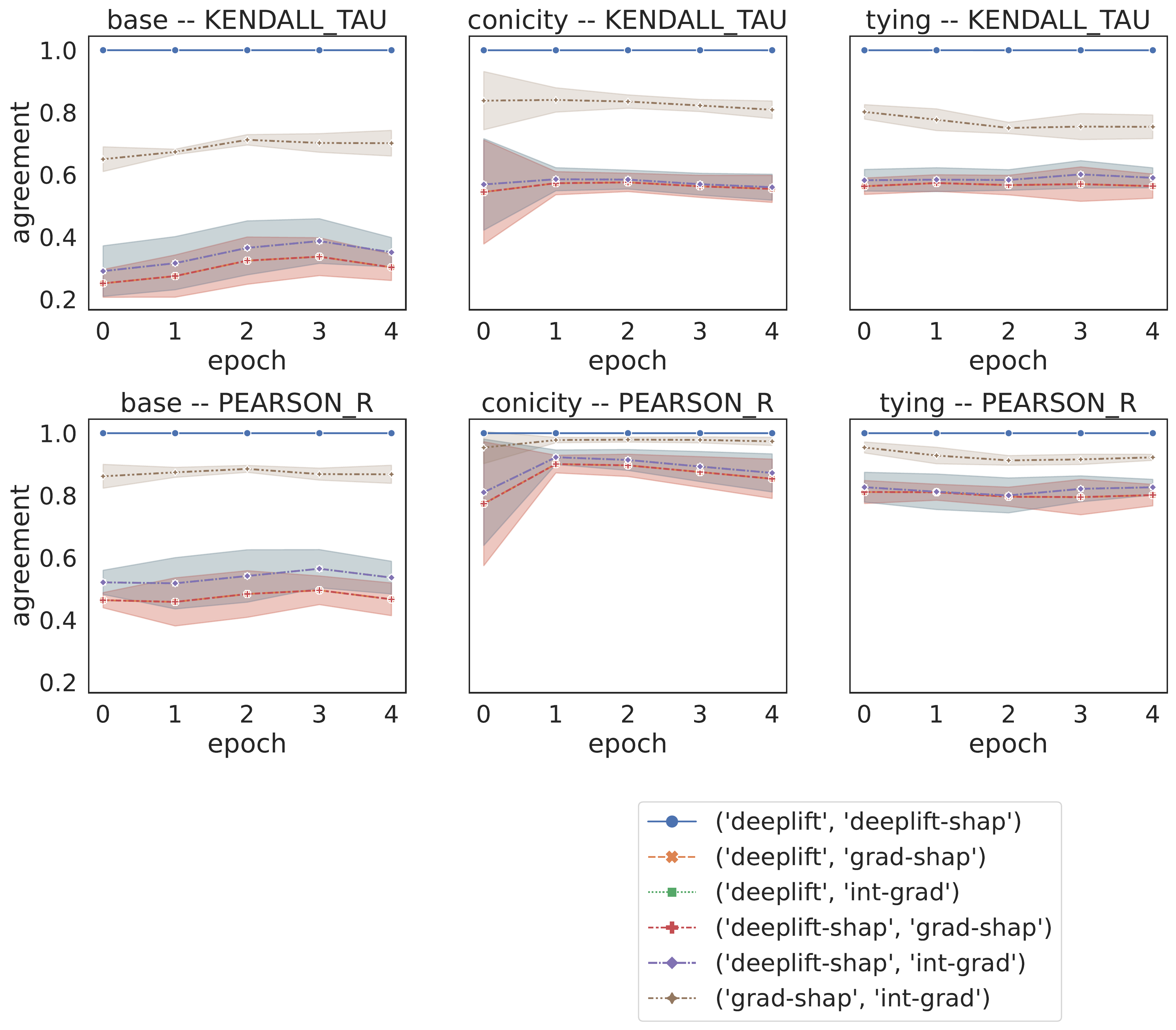}
\caption{\textsc{dbert} -- SST}
\label{fig:corr-dbert-subj}
\end{figure}

\begin{figure}
\centering
\includegraphics[width=\linewidth]{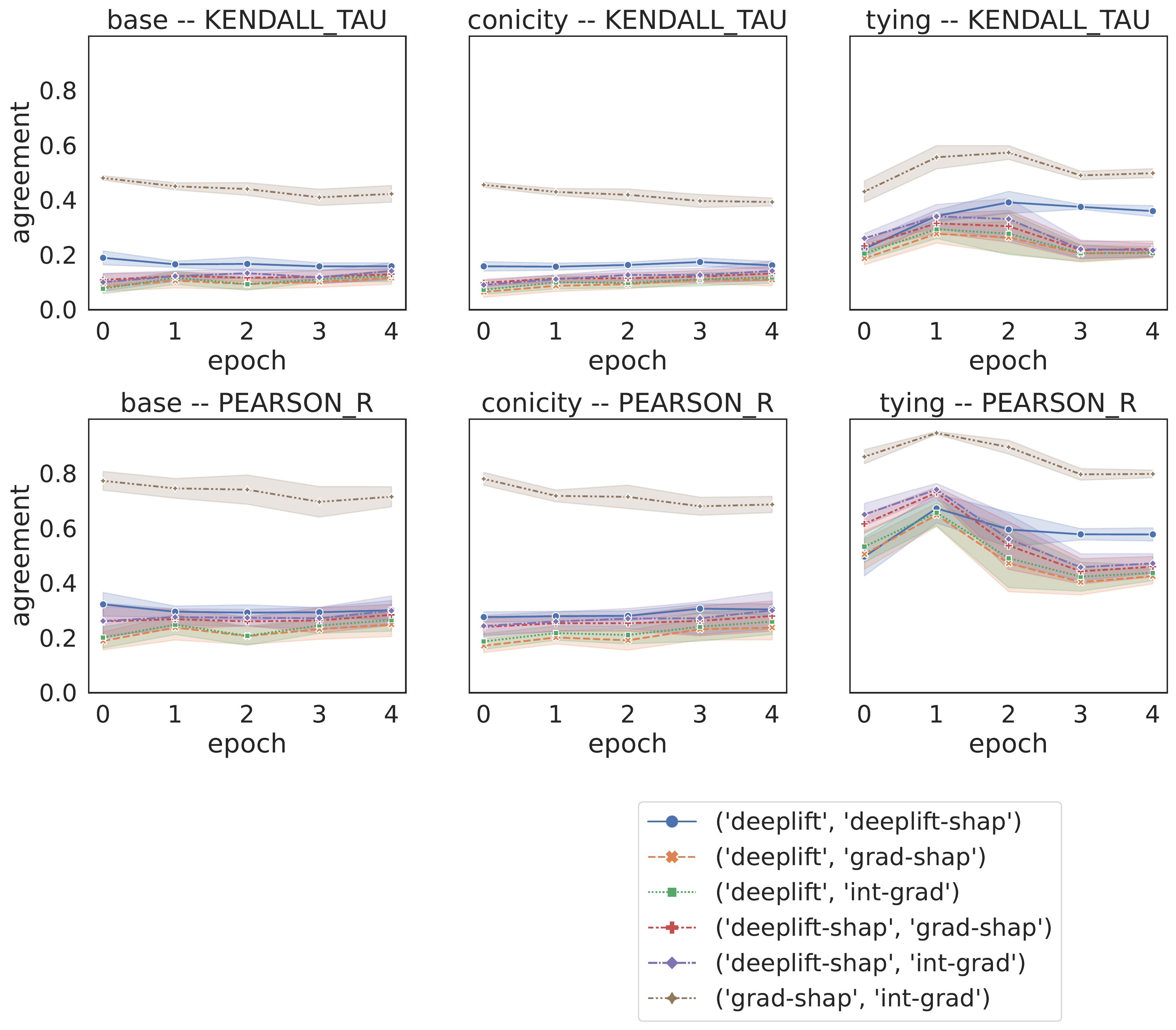}
\caption{\textsc{dbert} -- TREC}
\label{fig:corr-dbert-trec}
\end{figure}

\begin{figure}
\centering
\includegraphics[width=\linewidth]{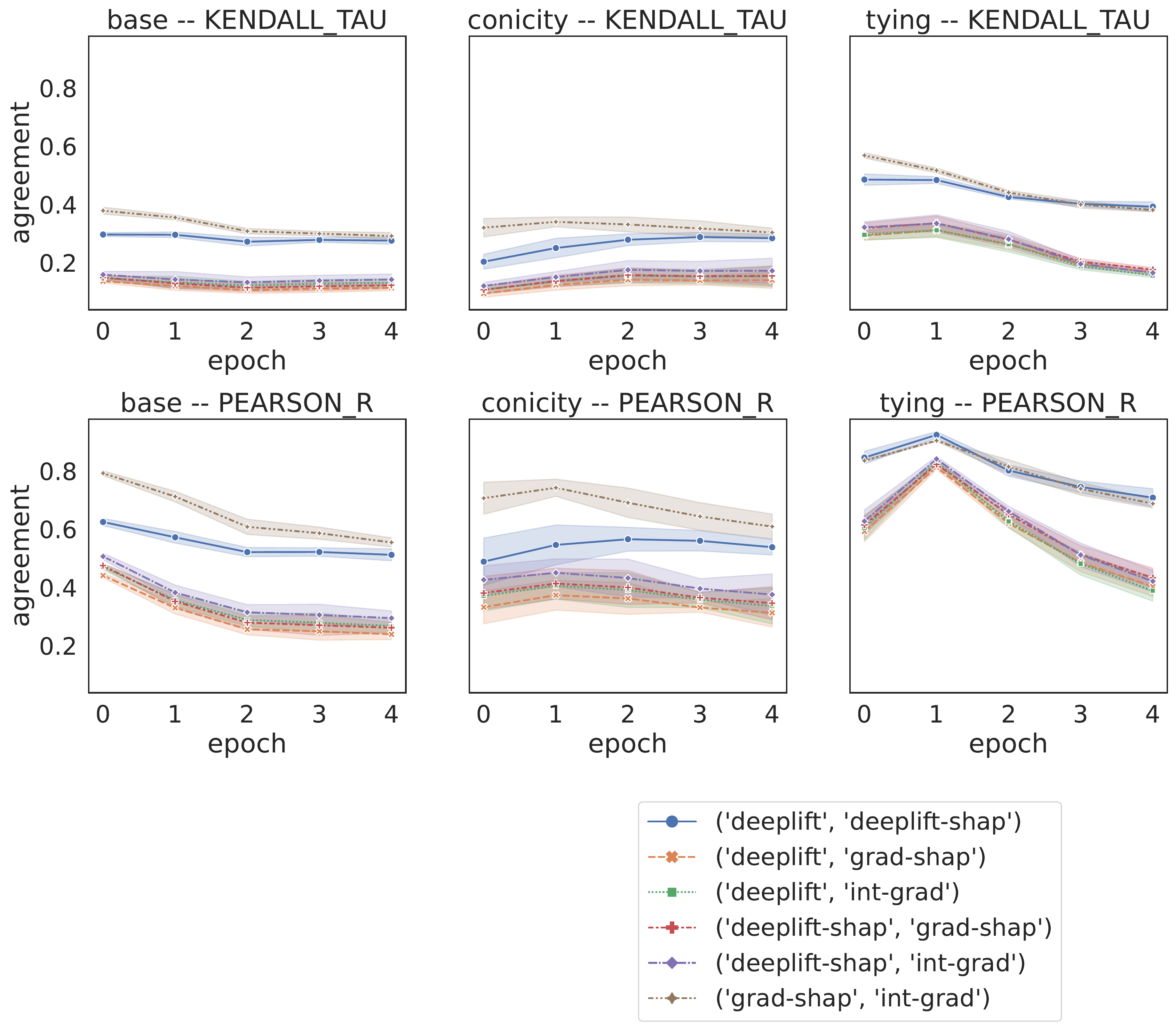}
\caption{\textsc{dbert} -- IMDB}
\label{fig:corr-dbert-imdb}
\end{figure}

We show the full version of local curvature statistics in \Cref{tab:repr_dbert_full} (without averaging over datasets). In \Cref{fig:corr-jwa-subj,fig:corr-jwa-sst,fig:corr-jwa-trec,fig:corr-jwa-imdb,fig:corr-dbert-subj,fig:corr-dbert-sst,fig:corr-dbert-trec,fig:corr-dbert-imdb} we plot correlation scores ($k_\tau$ and $p_r$) with standard deviation on the test splits. We include the results for all datasets across training epochs for regularized models (\textsc{conicity}, \textsc{tying}) when compared to their unregularized, \textsc{base} variants. 


\begin{table*}
\centering
\small
\begin{tabular}{lrrrrrrr}
\toprule
& & \multicolumn{2}{c}{Confidence} & \multicolumn{2}{c}{Ambiguity} & \multicolumn{2}{c}{Variability} \\
\cmidrule(lr){3-4} \cmidrule(lr){5-6} \cmidrule(lr){7-8}
& & B & T & B & T & B & T \\
\midrule
\multirow{4}{*}{\rotatebox[origin=c]{90}{Grad norm}}
& SUBJ & $-.37_{.00}$ & $-.06_{.00}$ & $.48_{.00}$ & $.05_{.02}$ & $.45_{.00}$ & $.00_{.83}$ \\
& SST & $-.40_{.00}$ & $.02_{.34}$ & $.58_{.00}$ & $.08_{.00}$ & $.42_{.00}$ & $-.23_{.00}$ \\
& TREC & $-.32_{.00}$ & $.05_{.32}$ & $.39_{.00}$ & $-.12_{.01}$ & $.38_{.00}$ & $.18_{.00}$ \\
& IMDB & $-.46_{.00}$ & $-.11_{.00}$ & $.61_{.00}$ & $.17_{.00}$ & $.60_{.00}$ & $.30_{.00}$ \\
\midrule
\multirow{4}{*}{\rotatebox[origin=c]{90}{Min dist}}
& SUBJ & $-.59_{.00}$ & $.01_{.70}$ & $.79_{.00}$ & $-.10_{.00}$ & $.74_{.00}$ & $.06_{.00}$ \\
& SST & $-.45_{.00}$ & $.30_{.00}$ & $.70_{.00}$ & $-.44_{.00}$ & $.49_{.00}$ & $.30_{.00}$ \\
& TREC & $-.55_{.00}$ & $.17_{.00}$ & $.70_{.00}$ & $-.26_{.00}$ & $.68_{.00}$ & $.25_{.00}$ \\
& IMDB & $-.53_{.00}$ & $.50_{.00}$ & $.70_{.00}$ & $-.74_{.00}$ & $.64_{.00}$ & $.04_{.00}$ \\
\bottomrule
\end{tabular}
\caption{Correlations between local curvature statistics in the representation space and cartography attributes for each dataset. We use average gradient norms (\textit{grad norm}) of the hidden representation with respect to the input embeddings and average distance to the nearest instance (\textit{min dist}). The columns correspond to the cartography attributes. We report the results for the unregularized model (B) and the regularized one to which we applied tying (T). The values in the subscript denote the standard deviation.}
\label{tab:repr_dbert_full}
\end{table*}

\end{document}